\documentclass{article} %
\usepackage{iclr2024_conference,times}
\usepackage{tabularx}
\usepackage[most]{tcolorbox}
\usepackage{wrapfig}
\usepackage{algorithm}
\usepackage{algpseudocode}
\usepackage{bbding}
\usepackage{graphicx}
\usepackage{subcaption}
\usepackage{enumitem}
\usepackage{pifont}
\usepackage{booktabs}
\usepackage{adjustbox}
\setlist[itemize]{leftmargin=10pt}

\usepackage{amsmath,amsfonts,bm}

\def\eqref#1{equation~\ref{#1}}

\def\1{\bm{1}}

\DeclareMathAlphabet{\mathsfit}{\encodingdefault}{\sfdefault}{m}{sl}
\SetMathAlphabet{\mathsfit}{bold}{\encodingdefault}{\sfdefault}{bx}{n}

    \newcolumntype{Y}{>{\centering\arraybackslash}X}
    \usepackage{hyperref}
    \usepackage{url}
    \tcbset{
      aibox/.style={
        width=474.18663pt,
        top=10pt,
        colback=white,
        colframe=black,
        colbacktitle=black,
        enhanced,
        center,
        attach boxed title to top left={yshift=-0.1in,xshift=0.15in},
        boxed title style={boxrule=0pt,colframe=white,},
      }
    }
    \newtcolorbox{AIbox}[2][]{aibox,title=#2,#1}

\title{Voila-A: Aligning Vision-Language Models with User's Gaze Attention}

\author{Kun Yan\textsuperscript{\rm 1},
Lei Ji\textsuperscript{\rm 2}, 
Zeyu Wang\textsuperscript{\rm 3},
Yuntao Wang\textsuperscript{\rm 3}, 
Nan Duan\textsuperscript{\rm 2},
Shuai Ma\textsuperscript{\rm 1} \\
\textsuperscript{\rm 1}Beihang University, \textsuperscript{\rm 2}Microsoft Research Asia, \textsuperscript{\rm 3}Tsinghua University \\
\texttt{\{kunyan, mashuai\}@buaa.edu.cn, \{leiji, nanduan\}@microsoft.com,} \\
\texttt{\{wang-zy23@mails.,yuntaowang@\}tsinghua.edu.cn} \\
}

\iclrfinalcopy %
\begin{document}

\maketitle

\begin{abstract}
In recent years, the integration of vision and language understanding has led to significant advancements in artificial intelligence, particularly through Vision-Language Models (VLMs). However, existing VLMs face challenges in handling real-world applications with complex scenes and multiple objects, as well as aligning their focus with the diverse attention patterns of human users. In this paper, we introduce gaze information, feasibly collected by AR or VR devices, as a proxy for human attention to guide VLMs and propose a novel approach, Voila-A, for gaze alignment to enhance the interpretability and effectiveness of these models in real-world applications.
First, we collect hundreds of minutes of gaze data to demonstrate that we can mimic human gaze modalities using localized narratives. We then design an automatic data annotation pipeline utilizing GPT-4 to generate the VOILA-COCO dataset. Additionally, we innovate the Voila Perceiver modules to integrate gaze information into VLMs while preserving their pretrained knowledge.
We evaluate Voila-A using a hold-out validation set and a newly collected VOILA-GAZE Testset, which features real-life scenarios captured with a gaze-tracking device. Our experimental results demonstrate that Voila-A significantly outperforms several baseline models. By aligning model attention with human gaze patterns, Voila-A paves the way for more intuitive, user-centric VLMs and fosters engaging human-AI interaction across a wide range of applications.
\end{abstract}
\section{Introduction}
The integration of vision and language understanding has witnessed significant advancements in recent years, particularly through the development of Vision-Language Models (VLMs). These models have demonstrated remarkable performance in various tasks, such as visual question answering, image captioning, and visual storytelling, among others. However, despite their impressive performance, current VLM systems often lack the ability to align their focus with that of a human user, which can lead to suboptimal performance and reduced user satisfaction. To address this challenge, we introduce Voila-A, a novel approach for aligning VLMs with a user's gaze attention, aiming to improve the interpretability and effectiveness of these models in real-world applications.

Recent research in multimodal vision and language tasks has leveraged multimodal large language models (MLLMs) to achieve superior performance \citep{liu2023visual,li2023blip,alayrac2022flamingo}. These models primarily focus on learning alignment between vision input and text tokens for LLMs or designing learnable interaction layers to attend the vision input to the frozen LLM layers. The importance of aligning AI with human attention has been highlighted in previous research, which demonstrates that incorporating visual attention can lead to improved user experience \citep{land2006eye-movements-and-actions,tanriverdi2000interacting-with-eye,Piening_2021looking-for-info}. Additionally, there has been growing interest in grounded MLLMs, which investigate the fine-grain grounding capability between region-text pairs instead of image-text pairs and further conduct dense regional prediction tasks \citep{zhong2022regionclip,jin2023grill,zhou2023regionblip}.

Visual regions can be represented in various ways, such as bounding boxes \citep{zhu2016visual7w,liu2017referring}, points \citep{mani2020point}, or traces \citep{pont2020connecting,yan2021control}. To input regional information into models, several methods have been proposed, including concatenating cropped image patches with the original text/image as model input \citep{zhang2023gpt4roi,bracha2023disclip}, using masks or Gaussian maps to emphasize areas of user interest \cite{lin2020interactive,lin2022multi}, or encoding points, boxes, or traces into positional encodings \citep{kirillov2023segment,voigtlaender2023connecting}. While bounding boxes and points have been widely used in VLMs, gaze offers a more natural way to represent visual regions and are most similar to human gaze. In this work, we propose using gaze as a more convenient and interactive way of representing visual regions, especially in augmented reality and virtual reality scenarios. Specifically for gaze, previous works have proposed gaze-directed visual grounding \citep{qian2023gvgnet} and eye-gaze-guided vision transformers \citep{ma2023eye}. However, these approaches have limitations in terms of scalability and flexibility. Despite these advancements, the integration of gaze information into large VLMs remains an open challenge. A key challenge in achieving this alignment lies in the integration of gaze information into VLMs while preserving the pretrained knowledge.

To tackle this issue, we demonstrate that mouse trace data can be a proxy for gaze behavior modeling and leverage trace data from Localized Narratives \citep{pont2020connecting} to annotate instructional data using GPT-4 \citep{openai2023gpt4}. We further design Voila-A's attention mechanism to incorporate gaze information while not forgetting pretrained knowledge. We evaluate Voila-A through a hold-out validation set and a newly collected test set of 100 samples, featuring real-life scenarios with a gaze-tracking device.

In this paper, we make the following contributions:

\begin{itemize}
\vspace{-5pt}
\item We propose Voila-A, a novel approach for aligning VLMs with a user's gaze attention, and design innovative mechanisms to integrate gaze information into VLMs while preserving pretrained knowledge. 
\item We leverage trace data from Localized Narratives to annotate instructional data using GPT-4, generating the VOILA-COCO dataset with \~72k QA pairs, and demonstrate the scalability of this method.
\item We evaluate Voila-A through a hold-out validation set and a newly collected VOILA-GAZE testset of 100 real gaze samples, demonstrating that our approach significantly outperforms several baselines, including Otter \citep{li2023otter} and Kosmos-2 \citep{peng2023kosmos}.
\vspace{-5pt}
\end{itemize}

Voila-A represents a significant step forward in the development of more intuitive and user-centric VLMs. By aligning model attention with human gaze patterns and leveraging state-of-the-art techniques, we pave the way for more effective and engaging human-AI interaction in a wide range of applications.

\section{Bridging the Gap in Daily Life Usage of Current VLMs through Gaze Integration}
\begin{figure}[t]
\vspace{-35pt}
\centering  {\includegraphics[width=1.0\linewidth]{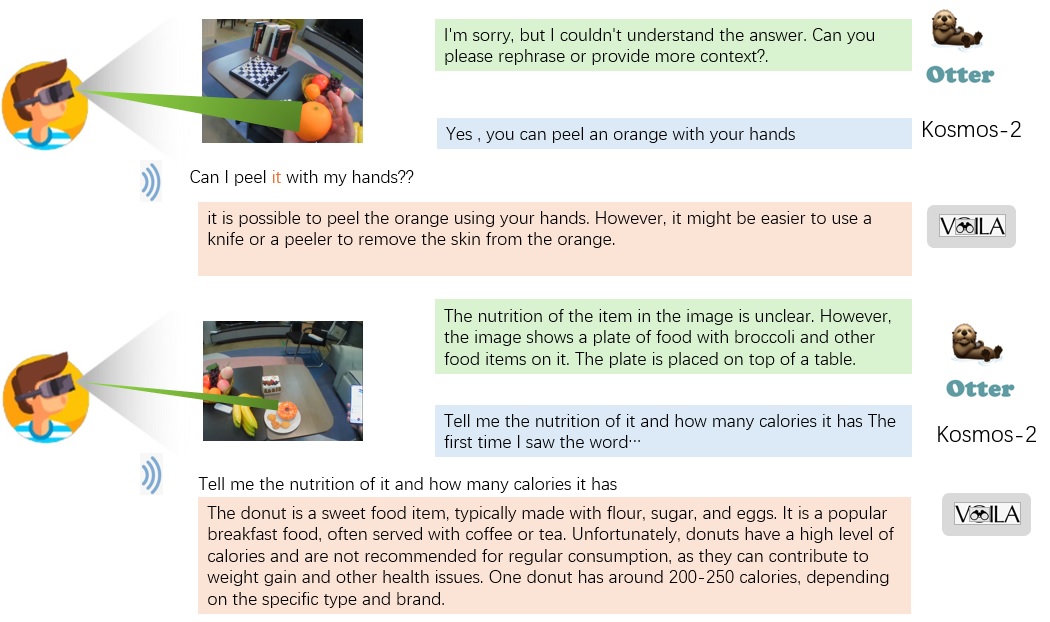}}    \caption{AR and VR scenarios usually involve complex scenes with multiple objects. Users may interested in only one specific object and gaze is the most natural way to interact with the device.
   }
   \label{fig:voila1}
\end{figure}
Although Vision-Language Models (VLMs) exhibit strong performance in various tasks, their applicability in everyday scenarios is hindered by their limited alignment with human users' focus. This misalignment leads to suboptimal performance and decreased user satisfaction. Current VLMs' inability to process these intentional modalities results in imprecise and unhelpful responses, as demonstrated in Figure~\ref{fig:voila1}. As shown in Figure~\ref{fig:voila}, a user's intent can be communicated through spoken language, multimodal expressions, or even be concealed. Gaze direction data can clarify vague expressions, while uncovering hidden intentions is more challenging.

{ \cite{Zhang2020HumanGA} provides an overview of gaze-related research, outlining a process that begins with collecting human gaze data (further discussed in \ref{rw-gazedata}), followed by building models to predict human attention distribution (i.e., saliency models, as discussed in \ref{rw-saliency}), and culminating in human-gaze-assisted AI. They acknowledge that \textit{\textbf{AI agents capable of perceiving and understanding human gaze behaviors can better infer user needs and assist in daily tasks}}. However, they also note that \textit{\textbf{research in this final direction is still limited}}. Our work aims to advance this area further. \
Incorporating gaze data into VLMs improves their applicability and effectiveness in everyday scenarios. This integration enables VLMs to focus on users' interests, delivering accurate and relevant responses while understanding intent, context, and preferences. As spatial computing advances, gaze data becomes essential for dynamic, interactive environments, allowing VLMs to adapt in real-time and offer intuitive, seamless user experiences.
}
\section{Leveraging Trace Data as an Alternative Approach to Align VLMs with Gaze Attention}

Obtaining gaze data for training VLMs can be challenging, as it is difficult to annotate and expensive to acquire. However, an alternative approach can be employed to align VLMs with user gaze attention: utilizing trace data, such as mouse traces, which we demonstrate to have similarities with gaze data. In this section, we discuss the potential of trace data as a proxy for gaze data and propose a method for transforming trace data to make it more gaze-like, ultimately enabling the effective use of trace data for aligning VLMs with user gaze attention.

Localized Narratives~\citep{pont2020connecting}, a prior work, have annotated 849,000 images with mouse traces that are aligned with each word of the descriptions. The project involved 156 professional annotators who worked full-time, with annotator managers ensuring high-quality annotations through manual inspections and an automatic quality control mechanism. After discarding 23.5\% of annotations, the remaining ones demonstrated a semantic accuracy of 98.0\% for nouns and verbs. The accuracy of mouse traces in relation to object locations was also analyzed, revealing that most trace points were within the correct bounding box.

By collecting hundreds of minutes of gaze data samples as described in \ref{datacollect}, we find that gaze and mouse traces exhibit similarities, as users tend to fix their gaze on the target object when asking questions, a behavior also observed with mouse traces. However, there are minor differences between the two, specifically in terms of gaze fixation continuity and the presence of noise points outside the target object at the end of a query. In the case of mouse traces, points that fell outside the bounding box were attributed to two factors: annotators often circled around objects, causing the traces to be near but not inside the box, and some annotators started moving the mouse before describing the object or vice versa. These observations provide valuable insights for properly leveraging trace data into the alignment process and understanding the relationship between gaze attention and language description.

In order to utilize trace data as a substitute for gaze data, we introduce a method to transform mouse traces, thereby reducing the discrepancies between the two data types and making the trace data more gaze-like. We first address the inherent noise in both trace points and gaze points by converting them into 2D heatmaps using Gaussian blur:

\begin{equation}
H(x, y) = \frac{1}{2\pi\sigma^2} e^{-\frac{x^2 + y^2}{2\sigma^2}}
\end{equation}

where $H(x, y)$ represents the heatmap value at position $(x, y)$, and $\sigma$ is the standard deviation of the Gaussian kernel.

Since mouse traces are more continuous than gaze fixations, we downsample the trace data to better resemble the distribution of gaze data. We investigate the Earth Mover's Distance (EMD) between the mean heatmaps of 1k gaze and trace samples while varying the sampling rate from 1 to 40:

\begin{equation}
\text{EMD}(P, Q) = \frac{\sum_{i=1}^{n} |F_i(P) - F_i(Q)|}{\sum_{i=1}^{n} F_i(P)}
\end{equation}

where $P$ and $Q$ are the distributions of the gaze and trace heatmaps, $F_i$ denotes the cumulative distribution function, and $n$ is the number of bins.

\begin{wrapfigure}[17]{h}{0.5\columnwidth}
    \centering
    \includegraphics[width=0.48\columnwidth]{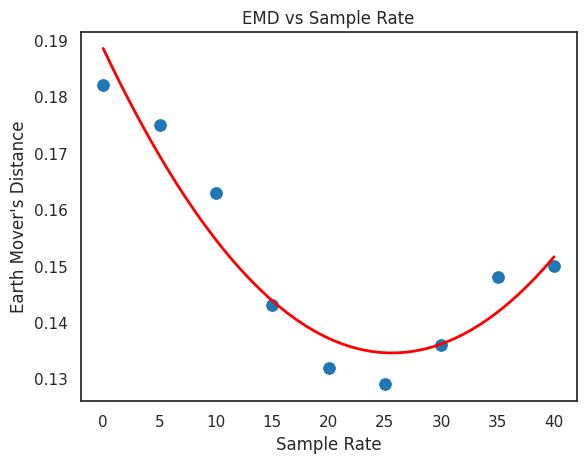}
    \caption{EMD between the mean heatmaps of 1k gaze and trace samples with varying sampling rates.}
    \label{fig:emd}
\end{wrapfigure}

We observe that the EMD has a local minimum value around a sampling rate of 25 as shown in Figure \ref{fig:emd}. By selecting this optimal sampling rate, we can approximate the trace heatmap as an alternative to the real gaze heatmap from a statistical perspective. Consequently, this transformation mitigates the differences in inter-relationships, compactness, and noisiness between the trace and gaze data.

\section{Method}
\subsection{Automatic Data Annotation For LN-COCO}

\begin{wrapfigure}[17]{ht}{0.5\columnwidth}
    \centering
    \includegraphics[width=0.48\columnwidth,clip, trim=6.5cm 4cm 3.9cm 3cm]{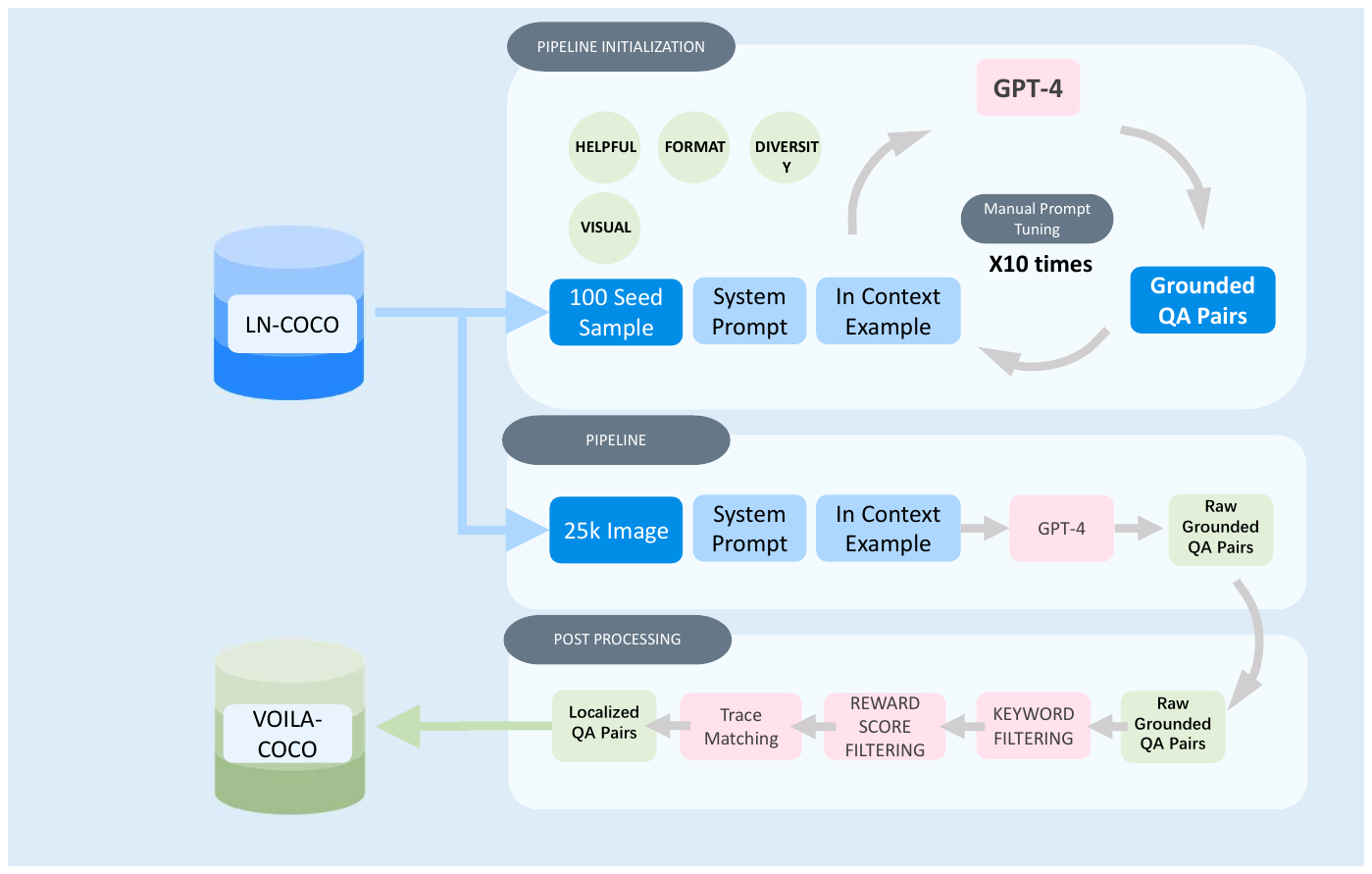}
    \caption{Automatic Data Annotation Pipeline}
    \label{fig:datapipe}
    \vspace{-25pt}
\end{wrapfigure}

The automatic data annotation process for Voila-A is driven by the motivation to develop a more intuitive and user-centric VLM by aligning model attention with human gaze patterns. As shown in Figure~\ref{fig:datapipe}, this process aims to create an effective and engaging human-AI interaction experience across various applications. To achieve this, we have designed an innovative prompting approach that leverages the capabilities of GPT-4 as a visual assistant to annotate trace-aligned instructional data to simulate the user's gaze attention. The data annotation process follows design principles to ensure accurate, relevant, and consistent annotations. These include: 1) focusing on referable sentences and appropriate tags, 2) using a conversational format with specific and general questions, 3) addressing various visual content aspects with definite answers, 4) incorporating complex questions while avoiding uncertainty, and 5) offering detailed, well-organized explanations.

As illustrated in Figure \ref{fig:datapipe}, the automatic data annotation pipeline comprises three stages. 

\textbf{Stage 1: Prompt Design Iteration.} The first stage focuses on refining the prompt design. Let $S = \{(I_i, N_i, T_i, C_i)\}_{i=1}^{100}$ be a set of 100 samples from the LN-COCO dataset, where $I_i$ represents the image, $N_i$ the localized narrative, $T_i$ the corresponding trace, and $C_i$ the set of five captions from COCO-caption. We initiate the process with a basic system prompt, instructing GPT-4 to generate direct questions $Q_{i,j}^D$ and indirect questions $Q_{i,j}^I$ and corresponding answers $A_{i,j}$ that specifically reference the localized narratives while considering COCO-caption as background information. The referring portions are annotated with a unique marker $\mathcal{M}$ for trace matching during post-processing. We also provide two in-context examples to guide the model in generating helpful, well-formatted, diverse, and visually grounded QA pairs. Throughout each iteration $k$, we manually evaluate the quality of the generated grounded QA pairs and adjust the prompt to enhance their helpfulness, formatting, diversity, and visual relevance. After $K = 10$ iterations, we find the quality of most pairs to be satisfactory, and subsequently, we freeze the prompt to initiate the pipeline.

\textbf{Stage 2: Data Sampling.} In the second stage, we sample $N = 25,000$ image pairs from the LN-COCO dataset and obtain approximately $M = 75,000$ QA pairs. 

\textbf{Stage 3: Post-processing.} The third stage involves post-processing the raw grounded QA pairs. This includes further filtering based on a set of keywords $\mathcal{K} = \{\text{"prompt", "this picture", "reference caption", ...}\}$. We define a filtering function $F_k(Q_{i,j}, A_{i,j}, \mathcal{K})$ that identifies and removes QA pairs containing meta descriptions of the prompt. We note that this issue may be further resolved by using GPT-4V, which was not available during our submission date. Additionally, we identify cases where answers are unhelpful, such as "I don't know" or "It's hard to tell." We find that these types of answers have low reward scores, so we further examine all pairs using a reward model~\citep{reward} and filter the dataset by setting a minimum reward threshold $\tau$. We define a filtering function $F_r(Q_{i,j}, A_{i,j}, \tau)$ that removes QA pairs with reward scores below $\tau$. Finally, we segment each localized narrative into temporally aligned segments with respect to the special marker $\mathcal{M}$. Each segment comprises a grounded fact, a corresponding trace, a direct and indirect question, and an answer. This forms the final VOILA-COCO dataset, denoted as $\mathcal{D} = \{(F_i, T_i, Q_{i,j}^D, Q_{i,j}^I, A_{i,j})\}$. It is worth noting that we did not utilize all localized narratives, leaving room for future exploration.We annotate the COCO subset of localized narratives, resulting in the Voila-COCO dataset, with statistics presented in Table~\ref{datastat-ln}.

The finalized prompt can be found in \ref{Prompt}. We also visualize a sample of our annotated data in Figure \ref{fig:voila-coco-example}. By adhering to these design principles, the automatic data annotation process ensures that the resulting dataset is of high quality and effectively aligns the VLM's attention with that of a human user.

\begin{table}[]
\caption{Statistics of Voila-COCO and Voila-Gaze Datasets}
\label{tab:statistics}
\centering
\begin{tabularx}{\textwidth}{l|c|c|c|X}
\hline
Dataset & Split & Number of Images & Number of Questions & Survival Rate from Raw Data \\ \hline
Voila-COCO & Training & 20000 & 70000 & 93.5\% \\
Voila-COCO & Validation & 100 & 550 & 71.1\% \\
Voila-COCO & Test & 500 & 1900 & 75.7\% \\
Voila-GAZE & Real-life & 100 & 100 & 22.2\% \\
\hline
\end{tabularx}
\label{datastat-ln}
\vspace{-15pt}
\end{table}

\subsection{VOILA-GAZE: Real-life gaze-QA pairs}
To further demonstrate the effectiveness of our method in aligning VLMs with real-life users' gaze attention, we conduct experiments in two everyday scenarios, encompassing a variety of question types details can be found in Table~\ref{tab:task_descriptions}.

In addition to the recorded gaze trajectory, video, and transcription, each participant is instructed to annotate the key elements of their questions, formulate clear questions based on their interpretations, and choose the best answer from three candidate answers generated by GPT-4 according to their annotations. The experiment includes 16 participants (8 per scenario) with an equal gender distribution, aged between 20 and 30 (with a standard deviation of 2.06). Each participant takes approximately 240 minutes to complete the study. After applying post-filtering and manual checking, we curate a set of 100 QA pairs as our real-life benchmark, VOILA-GAZE.The curation process is conducted by two individuals sequentially, with the second person double-checking the following aspects:
\textbf{1. The question is related and aligned with gaze.
2. The answer is meaningful and can be considered a proper response to the gaze-directed question.
3. The question is not related to specific brands, prices, or any other objects beyond general knowledge.
4. The question type is not biased towards a few simple patterns.}
This two-step process ensures the quality and relevance of the curated data while minimizing potential biases and maintaining a focus on general knowledge. Samples of VOILA-GAZE are shown in Figure~\ref{fig:voila-gaze-example}.

\subsection{Model Design}
\begin{figure}
    \centering
    \vspace{-5pt}
    \includegraphics[width=\linewidth,trim=1.6cm 5.6cm 0.9cm 5.4cm,clip]{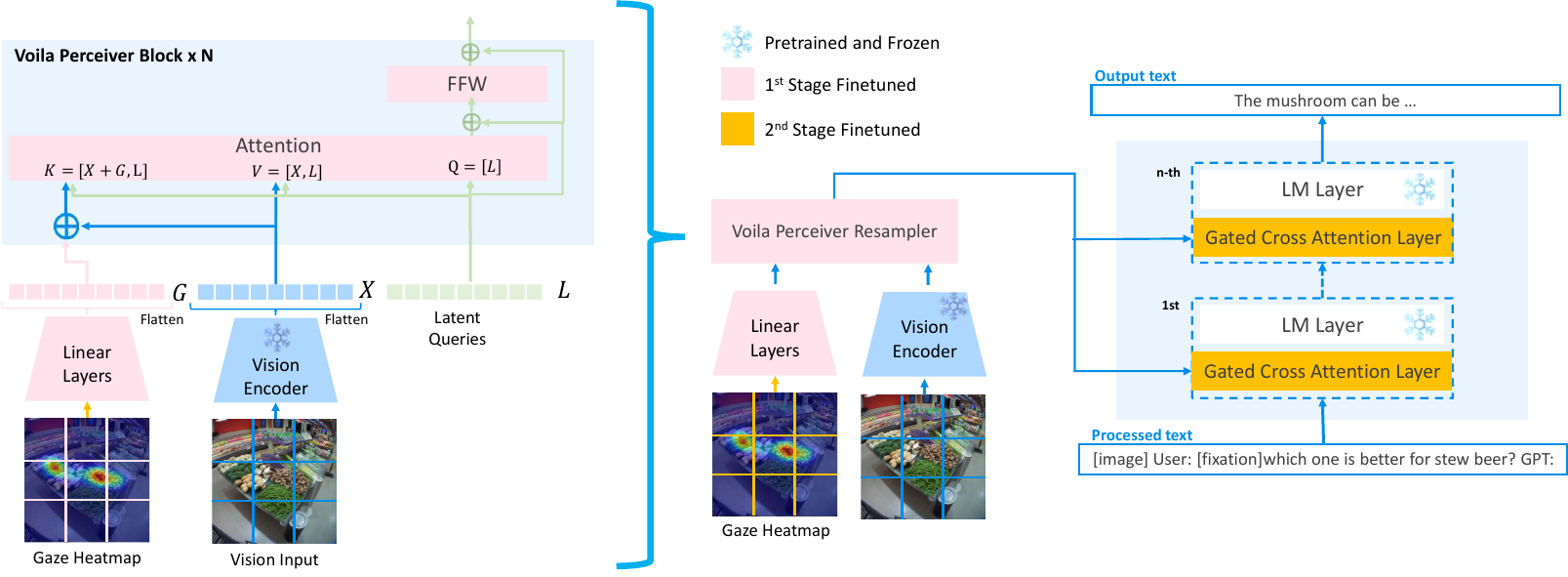}
    \caption{Overall Model Structure}
    \vspace{-20pt}
    \label{fig:gazepereiver}
\end{figure}

We employ the model architecture from OpenFlamingo, as illustrated on the right side of Figure \ref{fig:gazepereiver}. This framework consists of a pre-trained vision encoder, language decoder, and gated cross-attention layers, offering flexibility for multi-image and multi-turn conversational interactions. The primary challenge lies in incorporating gaze instructional signals into a pre-trained VLM. To tackle this issue, we initially develop several potential solutions, which are discussed in Sec~\ref{ablation-gaze}. Based on empirical evidence, we ultimately confirm the effectiveness of the Voila Perceiver Resampler solution. This approach comprises a series of Voila Perceiver Blocks (depicted on the left side of Figure \ref{fig:gazepereiver}). This innovative attention mechanism leverages gaze information to enhance visual feature perception. Our design adheres to the principle that gaze serves as an information aggregator in the attention process without disrupting the original learned distribution. The Voila Perceiver Block is defined as follows:
\begin{equation}
\begin{aligned}
    \textbf{VoilaPerceiverBlock}(x, L, G) = \text{LN}(L + \text{FF}(L + \text{Attn}(x, L, G)))
\end{aligned}
\end{equation}

where $x$ represents the image features, $G$ is the gaze heatmap patch encoded after simple linear layer. $L$ denotes the latent features. It is introduced from the original Perceiver as a small set of latent units that forms an attention bottleneck through which the inputs must pass.
The attention mechanism, $\text{Attn}(x, L, G)$, is computed as follows:
\begin{equation}
\begin{aligned}
    Q = \text{Linear}(L), \quad
    K = \text{Linear}(x \oplus L) + \text{Linear}(G \oplus P), \\
    V = \text{Linear}(x \oplus L), \quad
    \text{Attn}(x, L, G) = \text{Softmax}(Q K^\top) V
\end{aligned}
\end{equation}

Here, $\oplus$ denotes concatenation along the feature dimension, $P$ is zero padding with the same shape as $L$ and $\text{Linear}(\cdot)$ represents a linear transformation. The feed-forward network, $\text{FF}(\cdot)$, is a sequence of layer normalization, linear transformation, GELU activation, and another linear transformation.

The Voila Perceiver Resampler processes the input image features and gaze information, and then feeding them into series of Voila Perceiver Block:

\begin{equation}
\begin{aligned}
    \textbf{VoilaPerceiverResampler}(X, G) = \text{LN}(\text{Layer}(X, L, G))
\end{aligned}
\end{equation}

where $X$ denotes the input image features, $G$ represents the gaze information, and $\text{Layer}(X, L, G)$ is a sequence of Voila Perceiver Blocks. To obtain the gaze information $G$, we first divide the gaze heatmap $G'$ into patches. Then, we apply a linear transformation followed by layer normalization. The process can be represented by the following equation:
\begin{equation}
\begin{aligned}
    G = \text{LN}(\text{Linear}(G'))
\end{aligned}
\end{equation}
\subsection{Training}
Our approach utilizes the OpenFlamingo training paradigm to train the Voila model, building upon the pretrained weights of the Otter model, which incorporates an MPT-7B\citep{MosaicML2023Introducing} language encoder and a CLIP ViT-L/14~\citep{radford2021learning} vision encoder. To avoid overfitting and maximize the benefits of pretrained knowledge, we initially freeze both encoders. As shown in Figure~\ref{fig:gazepereiver}, we then train only the linear layers directly related to gaze input at the first stage for one epoch before fine-tuning the entire Perceiver resampler module, the cross-attention layers integrated into the language encoder, and the input/output embeddings of the language encoder in second stage for an additional epoch. This process results in roughly 1.3 billion trainable parameters for the Otter model.

During training, we adhere to a specific format for preparing our training data. This format combines an image, user instruction, "GPT"-generated answers 1, and a unique token known as the [endofchunk] token. We arrange the training data as follows:

\(<\)context\(>\) [image] User:[fixation]\(<\)instruction\(>\) GPT:[answers] \(<\)answer\(>\).[endofchunk]

Here, the [image], [answer], [fixation], and [endofchunk] tokens are distinct and serve particular functions. We adopt a chatbot-like format to enhance the instruction-following capabilities and conversational generalizability of our model. The [image] and [endofchunk] tokens originate from the OpenFlamingo training paradigm, while the [answer] token is a new addition by Otter. The [answer] token separates answers from instructions, allowing us to mask all tokens following the [answer] token during training and designate them as the model's prediction objectives. We also introduce the [fixation] token to direct the language model to utilize gaze information. We train our model using a cross-entropy loss function.

\section{Experiment}

\subsection{Evaluation metrics}
\paragraph{GPT-4 RANKING}
We utilize GPT-4 RANKING as our primary automated evaluation metric to assess model performance through a one-to-one comparison. The GPT Ranking represents the language model's evaluation of the quality of the generated response. This score signifies the extent to which the response aligns with the ground truth image description and answer while demonstrating the model's language proficiency. Factors such as grammar, semantics, and fluency are taken into account when comparing the response to that of another model.
\textbf{It is important to note that GPT-4 exhibits sequence ordering bias. }To mitigate this issue, we implement a dual-setting approach that reverses the order of the models, ensuring that the order does not influence the outcome. The prompt and evaluation procedure can be found in Figure \ref{GPT-RANK}.
\paragraph{Reward Score}
Given that our dataset is automatically annotated using GPT-4, it is crucial to mitigate any potential system bias during model evaluation. To this end, we incorporate human preference by utilizing a reward model score as an auxiliary metric. The reward model, which assesses the human-like quality of a response, is trained using human feedback to predict the superiority of a generated answer in relation to a given question from a human perspective\citep{reward}. This approach allows us to achieve a more balanced and robust evaluation process, ensuring that our model's performance aligns with human expectations and preferences.
\subsection{Main Results}
\subsubsection{VOILA Exhibits a Balanced Capability Between Helpfulness and Fact Grounding}
\begin{wrapfigure}[10]{t}{0.6\columnwidth}
\vspace{-10pt}
    \centering
    \includegraphics[width=0.33\linewidth]{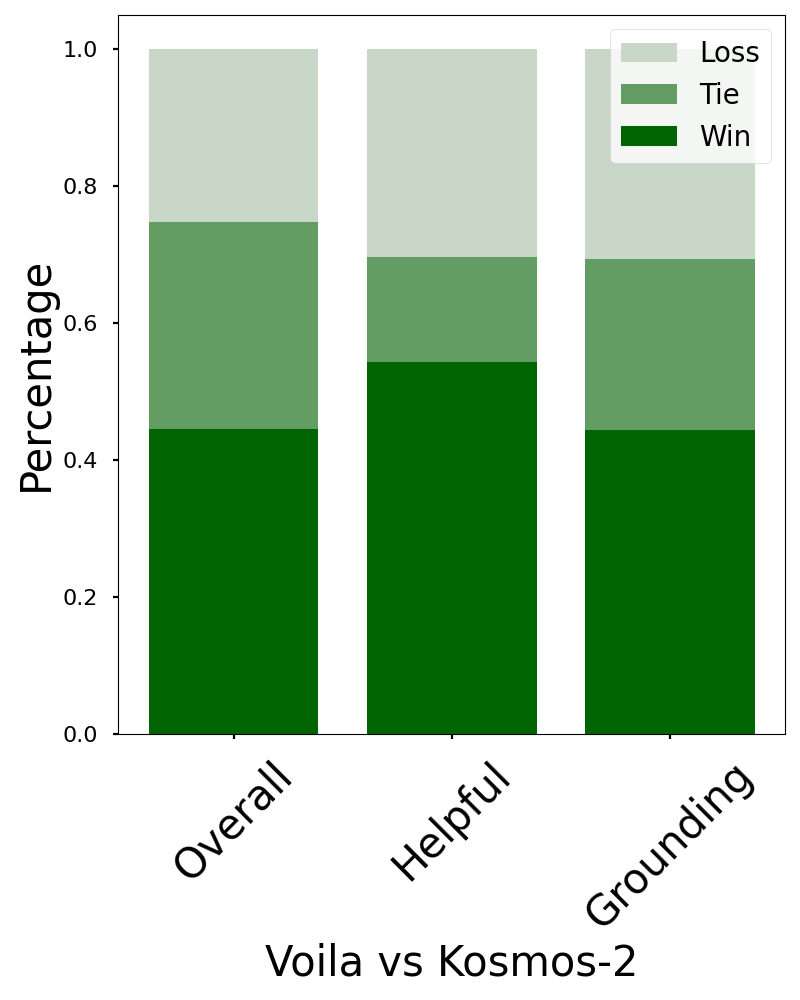}\includegraphics[width=0.33\linewidth]{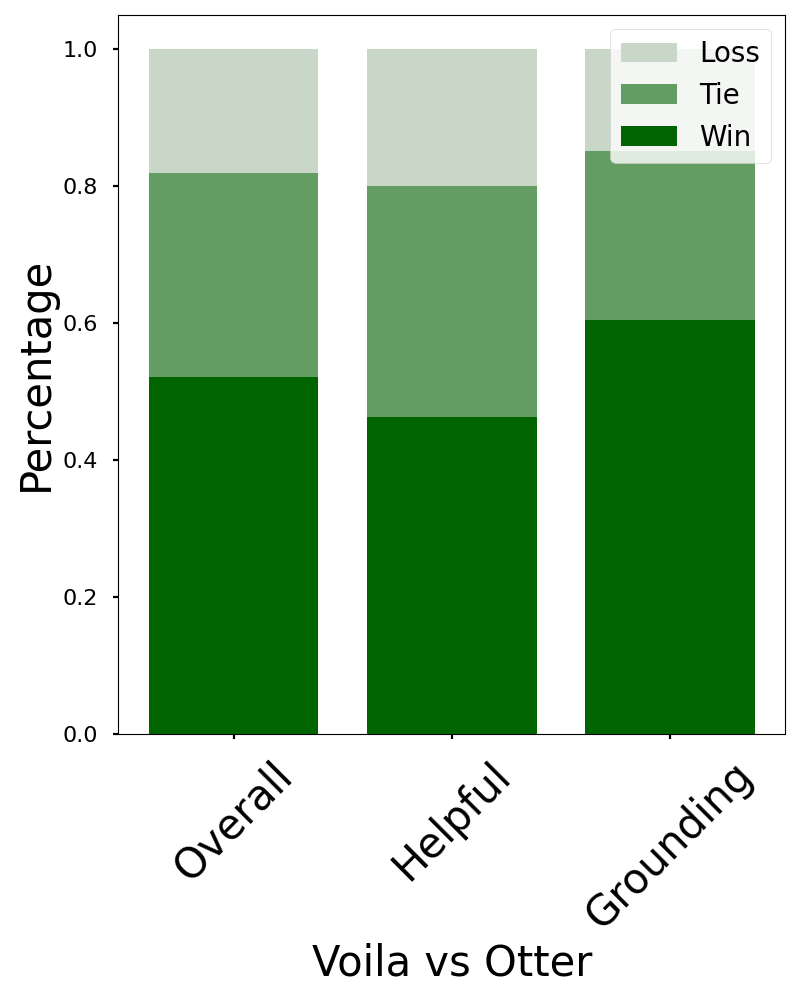}\includegraphics[width=0.33\linewidth]{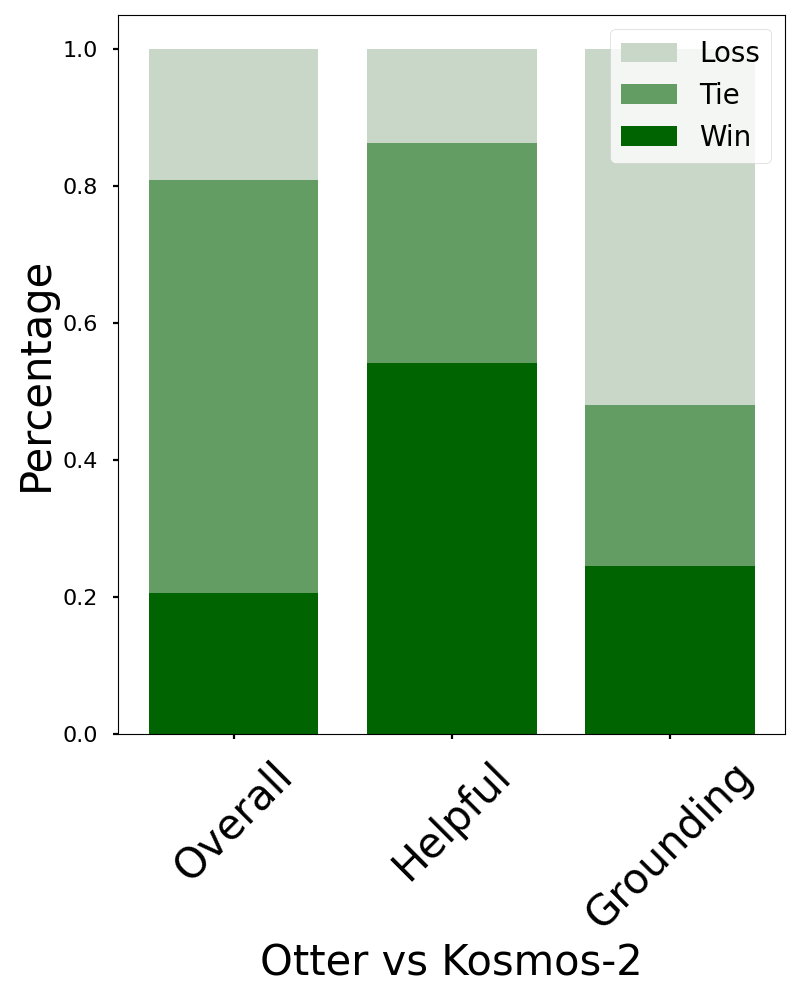}
    \caption{GPT-RANKING ON VOILA-COCO-Testset}
    \label{fig:coco-test}
\end{wrapfigure}

In Figure~\ref{fig:coco-test}, we observe a notable superiority of VOILA over both Otter and Kosmos-2 on the VOILA-COCO-TESTSET. Regarding the \emph{grounding} capability, both VOILA and Kosmos-2 trained with fine-grained grounded facts outperform Otter model in a large extent. Besides, VOILA surpasses Kosmos-2 marginally. With respect to \emph{helpful} capability, Otter delivers significantly more helpful responses than Kosmos-2. Since Otter is trained on top of Openflamingo with instruction-following dataset, it can provide more helpful response especially for informative queries while Kosmos-2 tends to answer visual observation from the input image. In addition, VOILA trained on gaze dataset demonstrates stronger helpful capabilities over all models. 
\begin{wrapfigure}[13]{H}{0.6\columnwidth}
    \centering
\includegraphics[width=0.33\linewidth]{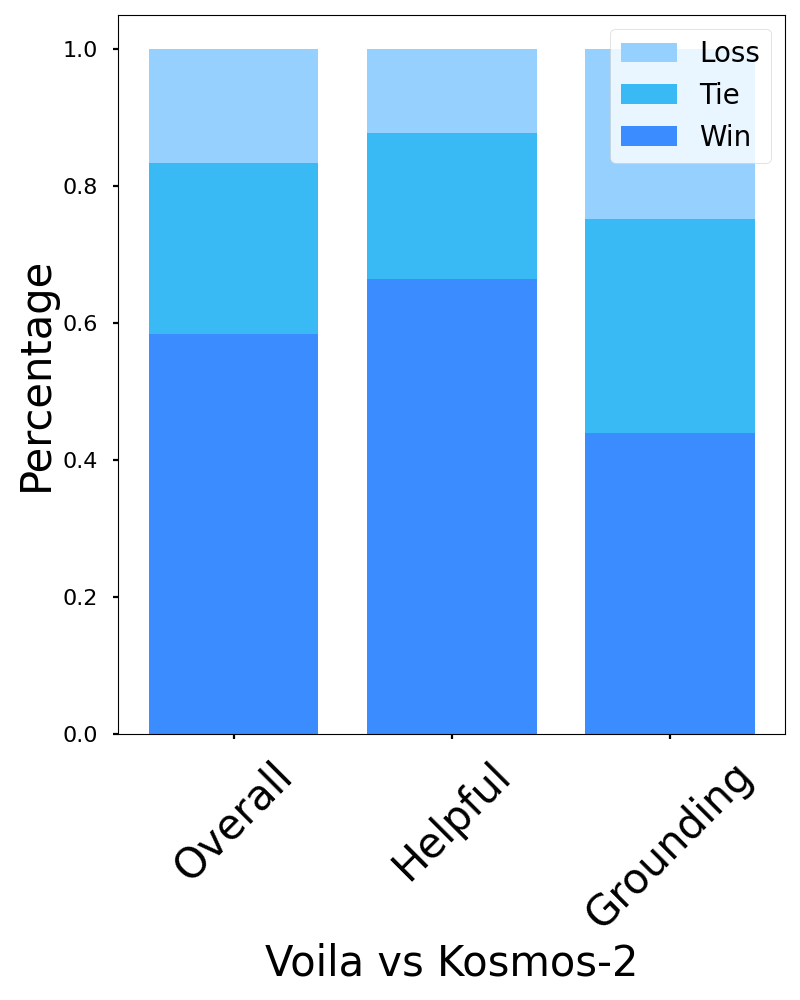}\includegraphics[width=0.33\linewidth]{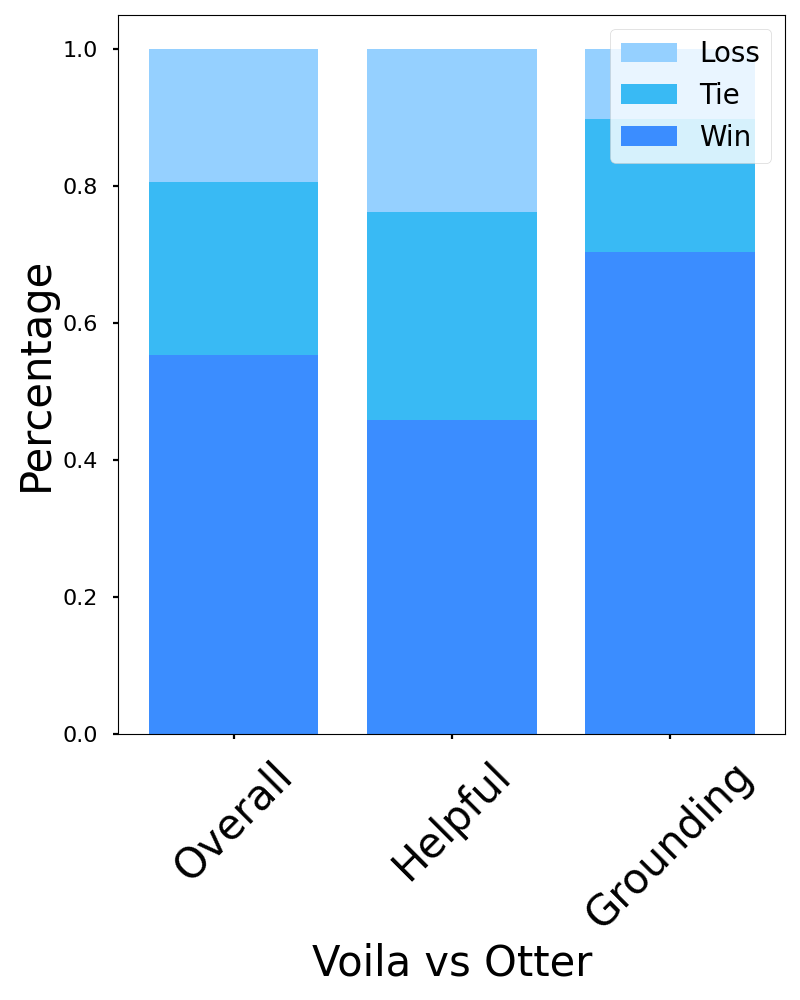}\includegraphics[width=0.33\linewidth]{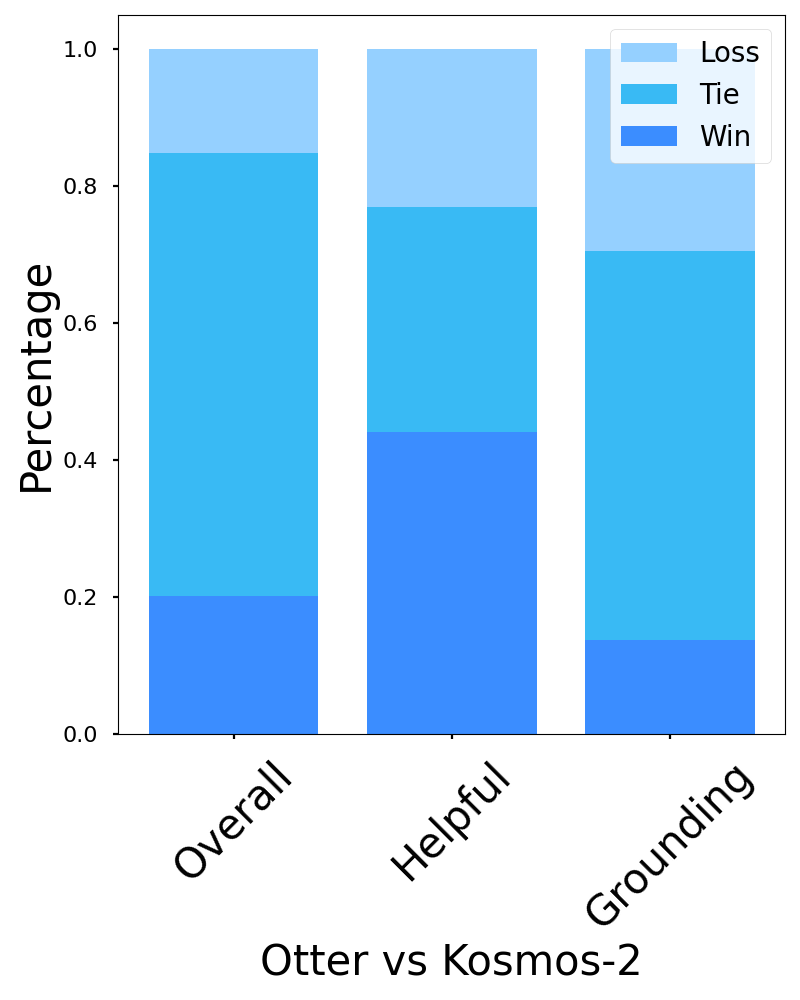}
    \caption{GPT-RANKING ON VOILA-GAZE}
    \label{fig:gaze-eval}
\end{wrapfigure}

In real gaze scenarios, as illustrated in Figure~\ref{fig:gaze-eval}, VOILA outperforms the two baseline models as well. These scenarios differ substantially from the collected COCO images and present more challenging questions, necessitating a higher degree of accurate intent understanding and reasoning. Especially from the comparison of Otter vs Kosmos-2, we found that there are much more Tie results due to the hardness of the real cases. Despite these increased demands, VOILA continues to surpass both models, further showcasing its balanced proficiency in both helpfulness and fact grounding.

\subsection{Ablation studies}
\subsubsection{Query types has a significant impact on Response Quality}
\begin{table}[h]
\caption{Ablation on query types}
\label{tab:result_question}
\centering
\small
{
\begin{tabular}{l|c|c|c}
\hline
Methods &Question types &Winning Rate Over Otter-base & Reward Score\\ \hline
Otter-base & coreference query&-&-1.91\\
Otter-base & direct query&0.51&0.02\\
Voila & coreference query&0.41&-0.79\\
Voila & direct query&0.62&0.14\\
Voila & in-context prompt + coreference query&0.46&-0.02\\
Voila & in-context prompt + direct query&0.77&0.20\\ \hline
\end{tabular}
}

\end{table}
Table~\ref{tab:result_question} investigates the varying performance of different question types, specifically direct and implicit/coreference queries. As the base model Openflamingo was pretrained on direct queries, both Otter and VOILA perform better in this category as expected. In addition, it is nature for human to communicate with correfernce queries. VOILA maintains strong performance when handling coreference queries with the gaze as a guidance while the Otter model decreases a lot. Furthermore, we append in-context QA pairs prior to the current query and observe that the examples further improves the quality of the responses. 
In real-life situations, multi-turn conversations are involved in most interactions with many coreference queries. The design of in-context prompt can assist VOILA to demonstrate a superior ability. This improvement is evident across both direct and coreference query types.
\subsubsection{Heatmap is a better way to incorporate gaze}

\begin{table}[h]
\caption{Ablation on Methods of Integrating Gaze Data}
\label{tab:result_gaze}
\centering
{
\begin{tabular}{l|c|c}
\hline
Methods & Wining Rate over Otter-base   & Reward Score \\ \hline   
Otter-base &-&-1.91\\ \hline
Gaze as discrete position tokens&0.19&-2.44 \\ 
Gaze \emph{bounding box} as image patch&0.36&-1.26\\  
Gaze \emph{bounding box} as discrete position tokens &0.21&-1.72\\ \hline
Voila(Gaze as heatmap)&0.41&-0.79  \\ 
\hline
\end{tabular}
}
\vspace{-10pt}
\end{table}
\label{ablation-gaze}
To establish the effectiveness of our approach, we implemented several alternative methods for incorporating gaze data into VLMs. These methods include: converting gaze sequences into discrete position tokens for LLMs, using the bounding box position of trace trajectories as additional patch tokens concatenated to VIT image feature token lists, and converting the bounding box coordinates into discrete tokens. { We provide an illustration of those method in Figure \ref{fig:model-design}. However, all these methods failed to outperform the gaze heatmap approach, as shown in Table~\ref{tab:result_gaze}. }

\subsubsection{Gradual Unfreezing of Parameters Yields Better Results}

\begin{table}[h]
\centering
{
\caption{Ablation on Training Procedure}
\label{tab:result_layers}
\begin{tabular}{l|c|c}
\hline
Layers fine-tuned &Winning Rate Over Otter-base & Reward Score \\ \hline  %
Otter-base frozen all&-&-1.91\\ 
Otter-base vision perceiver+cross attention&0.25&-1.78\\ 
Voila gaze weight&0.24&-1.52\\ 
Voila gaze weight+LORA&0.23&-1.02\\ 
Voila gaze weight\ding{221}perciever+cross attention&0.41&-0.79\\ \hline
\end{tabular}
}
\end{table}
Table~\ref{tab:result_layers} presents empirical findings that demonstrate the effectiveness of gradually unfreezing model parameters. Instead of directly unfreezing the vision perceiver and cross-attention layers, or using LORA to fine-tune the entire model, we first fine-tune the gaze-related weights and then fine-tune the other parts of the perceiver and cross-attention layers, which yields better results. We hypothesize that this improvement is due to the newly added gaze component needing to adapt to the distribution of the pretrained layers first. This adaptation process can further help mitigate the issue of catastrophic forgetting.

\subsection{Qualitative studies}

We conduct qualitative studies on randomly selected cases and demonstrate the results of several representative examples in Appendix Figure \ref{fig:casestudy}.{ According to the analysis, the conclusions can be summarized as: 1) Existing models are able to generate reasonable results for \textbf{explicit} queries. In the 1st row, the object \emph{cakes} and the attributes \emph{color} are explicitly mentioned in the query, and the three models are able to answer (partially) correctly; 2) Regarding to \textbf{coreference} queries, the model Otter is hard to understand the pronouns like \emph{it} without spatial guidance as shown in the 2nd row. This requires further context or generates the answer based on the salient object like \emph{plane} instead of the actual human attention; 3) The Kosmos-2 model can take the \textbf{bounding box} for grounding as spatial guidance, it is sometimes not accurate compared to the heatmap used in VOILA. As shown in the 3rd row, the bounding box is too coarse and made the model focus on the object \emph{plane} instead of the actual human attention \emph{sky}; 4) Besides, we found that Kosmos-2 tends to describe the detailed visual content and sometimes lacks the \textbf{instruction-following} capability; In the 4th row, the Kosmos-2 responses \emph{Keyboard} depicated in the bounding box ignoring the actual query intention; Finally, 5) There are still further challenges for all models to deal with. For instance, counting for objects requires intensive fine-grained recognition of the visual content demonstrated in the last row. }

\vspace{-15pt}
\section{Conclusion}
{In this study, we presented Voila-A, a cutting-edge approach that aligns Vision-Language Models with user gaze attention, thereby improving efficacy in practical applications. Voila-A can be implemented in HMD AR/VR devices as an egoview copilot, benefiting a wide range of users, including visually impaired individuals who rely on their gaze to communicate their intent. This method surpasses the capabilities of similar mobile apps that necessitate users to lift their phones for scene capture.
Despite its promising potential, there are limitations and future directions to consider. These include enhancing inference efficiency for real-time responses, integrating voice modalities for seamless interaction, and supporting higher resolutions for OCR and screen or UI understanding. We successfully utilized trace data to create the VOILA-COCO dataset, showcasing Voila-A's superior performance two benchmarks. Our research lays the foundation for more engaging human-AI interactions and encourages further exploration of Voila-A's integration with various modalities and tasks in the realm of multimodal AI systems.}
{
\section{Reproducibility Statement}
In the interest of promoting transparency and facilitating further research within the community, we are committed to providing comprehensive resources alongside the publication of our work. To this end, we will release the VOILA-COCO and VOILA-GAZE datasets, as well as the annotation pipeline, training code, and model weights. By making these materials publicly available, we aim to ensure that our methods can be easily replicated and built upon by other researchers. Our commitment to reproducibility aligns with the broader scientific goal of fostering collaboration and the development of novel ideas and techniques in the field.
}
\bibliography{iclr2024_conference}

\begin{thebibliography}{61}
\providecommand{\natexlab}[1]{#1}
\providecommand{\url}[1]{\texttt{#1}}
\expandafter\ifx\csname urlstyle\endcsname\relax
  \providecommand{\doi}[1]{doi: #1}\else
  \providecommand{\doi}{doi: \begingroup \urlstyle{rm}\Url}\fi

\bibitem[rew(2023)]{reward}
Openassistant/reward-model-deberta-v3-large-v2.
\newblock \url{https://huggingface.co/OpenAssistant/reward-model-deberta-v3-large-v2}, 2023.

\bibitem[Alayrac et~al.(2022)Alayrac, Donahue, Luc, Miech, Barr, Hasson, Lenc, Mensch, Millican, Reynolds, et~al.]{alayrac2022flamingo}
Jean-Baptiste Alayrac, Jeff Donahue, Pauline Luc, Antoine Miech, Iain Barr, Yana Hasson, Karel Lenc, Arthur Mensch, Katherine Millican, Malcolm Reynolds, et~al.
\newblock Flamingo: a visual language model for few-shot learning.
\newblock \emph{Advances in Neural Information Processing Systems}, 35:\penalty0 23716--23736, 2022.

\bibitem[Awadalla et~al.(2023)Awadalla, Gao, Gardner, Hessel, Hanafy, Zhu, Marathe, Bitton, Gadre, Sagawa, et~al.]{awadalla2023openflamingo}
Anas Awadalla, Irena Gao, Josh Gardner, Jack Hessel, Yusuf Hanafy, Wanrong Zhu, Kalyani Marathe, Yonatan Bitton, Samir Gadre, Shiori Sagawa, et~al.
\newblock Openflamingo: An open-source framework for training large autoregressive vision-language models.
\newblock \emph{arXiv preprint arXiv:2308.01390}, 2023.

\bibitem[Bracha et~al.(2023)Bracha, Shaar, Shamsian, Fetaya, and Chechik]{bracha2023disclip}
Lior Bracha, Eitan Shaar, Aviv Shamsian, Ethan Fetaya, and Gal Chechik.
\newblock Disclip: Open-vocabulary referring expression generation.
\newblock \emph{arXiv preprint arXiv:2305.19108}, 2023.

\bibitem[Chen et~al.(2023)Chen, Zhang, Zeng, Zhang, Zhu, and Zhao]{chen2023shikra}
Keqin Chen, Zhao Zhang, Weili Zeng, Richong Zhang, Feng Zhu, and Rui Zhao.
\newblock Shikra: Unleashing multimodal llm's referential dialogue magic.
\newblock \emph{arXiv preprint arXiv:2306.15195}, 2023.

\bibitem[Chen et~al.(2001)Chen, Anderson, and Sohn]{Chen2001WhatCA}
Mon-Chu Chen, John~R. Anderson, and Myeong-Ho Sohn.
\newblock What can a mouse cursor tell us more?: correlation of eye/mouse movements on web browsing.
\newblock \emph{CHI '01 Extended Abstracts on Human Factors in Computing Systems}, 2001.
\newblock URL \url{https://api.semanticscholar.org/CorpusID:16969703}.

\bibitem[Cho et~al.(2021)Cho, Lei, Tan, and Bansal]{cho2021unifying}
Jaemin Cho, Jie Lei, Hao Tan, and Mohit Bansal.
\newblock Unifying vision-and-language tasks via text generation.
\newblock In \emph{International Conference on Machine Learning}, pp.\  1931--1942. PMLR, 2021.

\bibitem[Dai et~al.(2023)Dai, Li, Li, Tiong, Zhao, Wang, Li, Fung, and Hoi]{dai2023instructblip}
Wenliang Dai, Junnan Li, Dongxu Li, Anthony Meng~Huat Tiong, Junqi Zhao, Weisheng Wang, Boyang Li, Pascale Fung, and Steven Hoi.
\newblock Instructblip: Towards general-purpose vision-language models with instruction tuning, 2023.

\bibitem[Das et~al.(2016)Das, Agrawal, Zitnick, Parikh, and Batra]{Das2016HumanAI}
Abhishek Das, Harsh Agrawal, C.~Lawrence Zitnick, Devi Parikh, and Dhruv Batra.
\newblock Human attention in visual question answering: Do humans and deep networks look at the same regions?
\newblock \emph{ArXiv}, abs/1606.03556, 2016.
\newblock URL \url{https://api.semanticscholar.org/CorpusID:220553}.

\bibitem[Gong et~al.(2023)Gong, Lyu, Zhang, Wang, Zheng, Zhao, Liu, Zhang, Luo, and Chen]{gong2023multimodal}
Tao Gong, Chengqi Lyu, Shilong Zhang, Yudong Wang, Miao Zheng, Qian Zhao, Kuikun Liu, Wenwei Zhang, Ping Luo, and Kai Chen.
\newblock Multimodal-gpt: A vision and language model for dialogue with humans.
\newblock \emph{arXiv preprint arXiv:2305.04790}, 2023.

\bibitem[Guo \& Agichtein(2010)Guo and Agichtein]{Guo2010TowardsPW}
Qi~Guo and Eugene Agichtein.
\newblock Towards predicting web searcher gaze position from mouse movements.
\newblock \emph{CHI '10 Extended Abstracts on Human Factors in Computing Systems}, 2010.
\newblock URL \url{https://api.semanticscholar.org/CorpusID:16330552}.

\bibitem[Hu et~al.(2021)Hu, Shen, Wallis, Allen-Zhu, Li, Wang, Wang, and Chen]{hu2021lora}
Edward~J Hu, Yelong Shen, Phillip Wallis, Zeyuan Allen-Zhu, Yuanzhi Li, Shean Wang, Lu~Wang, and Weizhu Chen.
\newblock Lora: Low-rank adaptation of large language models.
\newblock \emph{arXiv preprint arXiv:2106.09685}, 2021.

\bibitem[Huang et~al.(2011)Huang, White, and Dumais]{Huang2011NoCN}
Jeff Huang, Ryen~W. White, and Susan~T. Dumais.
\newblock No clicks, no problem: using cursor movements to understand and improve search.
\newblock \emph{Proceedings of the SIGCHI Conference on Human Factors in Computing Systems}, 2011.
\newblock URL \url{https://api.semanticscholar.org/CorpusID:882269}.

\bibitem[Huang et~al.(2012)Huang, White, and Buscher]{Huang2012UserSU}
Jeff Huang, Ryen~W. White, and Georg Buscher.
\newblock User see, user point: gaze and cursor alignment in web search.
\newblock \emph{Proceedings of the SIGCHI Conference on Human Factors in Computing Systems}, 2012.
\newblock URL \url{https://api.semanticscholar.org/CorpusID:15422846}.

\bibitem[Jiang et~al.(2015)Jiang, Huang, Duan, and Zhao]{jiang2015salicon}
Ming Jiang, Shengsheng Huang, Juanyong Duan, and Qi~Zhao.
\newblock Salicon: Saliency in context.
\newblock In \emph{The IEEE Conference on Computer Vision and Pattern Recognition (CVPR)}, June 2015.

\bibitem[Jin et~al.(2023)Jin, Mukherjee, Cheng, Shen, Chen, Awadallah, Jose, and Ren]{jin2023grill}
Woojeong Jin, Subhabrata Mukherjee, Yu~Cheng, Yelong Shen, Weizhu Chen, Ahmed~Hassan Awadallah, Damien Jose, and Xiang Ren.
\newblock Grill: Grounded vision-language pre-training via aligning text and image regions.
\newblock \emph{arXiv preprint arXiv:2305.14676}, 2023.

\bibitem[Judd et~al.(2009)Judd, Ehinger, Durand, and Torralba]{Judd2009LearningTP}
Tilke Judd, Krista~A. Ehinger, Fr{\'e}do Durand, and Antonio Torralba.
\newblock Learning to predict where humans look.
\newblock \emph{2009 IEEE 12th International Conference on Computer Vision}, pp.\  2106--2113, 2009.
\newblock URL \url{https://api.semanticscholar.org/CorpusID:16445820}.

\bibitem[Kienzle et~al.(2006)Kienzle, Wichmann, Scholkopf, and Franz]{Kienzle2006ANA}
Wolfgang Kienzle, Felix Wichmann, Bernhard Scholkopf, and Matthias~O. Franz.
\newblock A nonparametric approach to bottom-up visual saliency.
\newblock In \emph{Neural Information Processing Systems}, 2006.
\newblock URL \url{https://api.semanticscholar.org/CorpusID:11543059}.

\bibitem[Kim et~al.(2017)Kim, Bylinskii, Borkin, Gajos, Oliva, Durand, and Pfister]{kim2017bubbleview}
Nam~Wook Kim, Zoya Bylinskii, Michelle~A Borkin, Krzysztof~Z Gajos, Aude Oliva, Fredo Durand, and Hanspeter Pfister.
\newblock Bubbleview: an interface for crowdsourcing image importance maps and tracking visual attention.
\newblock \emph{ACM Transactions on Computer-Human Interaction (TOCHI)}, 24\penalty0 (5):\penalty0 36, 2017.
\newblock \doi{10.1145/3131275}.

\bibitem[Kingma \& Ba(2017)Kingma and Ba]{kingma2017adam}
Diederik~P. Kingma and Jimmy Ba.
\newblock Adam: A method for stochastic optimization, 2017.

\bibitem[Kirillov et~al.(2023)Kirillov, Mintun, Ravi, Mao, Rolland, Gustafson, Xiao, Whitehead, Berg, Lo, et~al.]{kirillov2023segment}
Alexander Kirillov, Eric Mintun, Nikhila Ravi, Hanzi Mao, Chloe Rolland, Laura Gustafson, Tete Xiao, Spencer Whitehead, Alexander~C Berg, Wan-Yen Lo, et~al.
\newblock Segment anything.
\newblock \emph{arXiv preprint arXiv:2304.02643}, 2023.

\bibitem[Koch \& Ullman(1985)Koch and Ullman]{Koch1985ShiftsIS}
Christof Koch and Shimon Ullman.
\newblock Shifts in selective visual attention: towards the underlying neural circuitry.
\newblock \emph{Human neurobiology}, 4 4:\penalty0 219--27, 1985.
\newblock URL \url{https://api.semanticscholar.org/CorpusID:45203429}.

\bibitem[K{\"u}mmerer et~al.(2016)K{\"u}mmerer, Wallis, and Bethge]{Kmmerer2016DeepGazeIR}
Matthias K{\"u}mmerer, Thomas S.~A. Wallis, and Matthias Bethge.
\newblock Deepgaze ii: Reading fixations from deep features trained on object recognition.
\newblock \emph{ArXiv}, abs/1610.01563, 2016.
\newblock URL \url{https://api.semanticscholar.org/CorpusID:13563819}.

\bibitem[Land(2006)]{land2006eye-movements-and-actions}
Michael~F Land.
\newblock Eye movements and the control of actions in everyday life.
\newblock \emph{Progress in retinal and eye research}, 25\penalty0 (3):\penalty0 296--324, 2006.

\bibitem[Li et~al.(2023{\natexlab{a}})Li, Zhang, Chen, Wang, Yang, and Liu]{li2023otter}
Bo~Li, Yuanhan Zhang, Liangyu Chen, Jinghao Wang, Jingkang Yang, and Ziwei Liu.
\newblock Otter: A multi-modal model with in-context instruction tuning.
\newblock \emph{arXiv preprint arXiv:2305.03726}, 2023{\natexlab{a}}.

\bibitem[Li et~al.(2023{\natexlab{b}})Li, Li, Savarese, and Hoi]{li2023blip}
Junnan Li, Dongxu Li, Silvio Savarese, and Steven Hoi.
\newblock Blip-2: Bootstrapping language-image pre-training with frozen image encoders and large language models.
\newblock \emph{arXiv preprint arXiv:2301.12597}, 2023{\natexlab{b}}.

\bibitem[Lin et~al.(2020)Lin, Zhang, Chen, Cheng, and Lu]{lin2020interactive}
Zheng Lin, Zhao Zhang, Lin-Zhuo Chen, Ming-Ming Cheng, and Shao-Ping Lu.
\newblock Interactive image segmentation with first click attention.
\newblock In \emph{Proceedings of the IEEE/CVF conference on computer vision and pattern recognition}, pp.\  13339--13348, 2020.

\bibitem[Lin et~al.(2022)Lin, Zhang, Han, and Lu]{lin2022multi}
Zheng Lin, Zhao Zhang, Ling-Hao Han, and Shao-Ping Lu.
\newblock Multi-mode interactive image segmentation.
\newblock In \emph{Proceedings of the 30th ACM International Conference on Multimedia}, pp.\  905--914, 2022.

\bibitem[Liu et~al.(2023)Liu, Li, Wu, and Lee]{liu2023visual}
Haotian Liu, Chunyuan Li, Qingyang Wu, and Yong~Jae Lee.
\newblock Visual instruction tuning.
\newblock \emph{arXiv preprint arXiv:2304.08485}, 2023.

\bibitem[Liu et~al.(2017)Liu, Wang, and Yang]{liu2017referring}
Jingyu Liu, Liang Wang, and Ming-Hsuan Yang.
\newblock Referring expression generation and comprehension via attributes.
\newblock In \emph{Proceedings of the IEEE International Conference on Computer Vision}, pp.\  4856--4864, 2017.

\bibitem[Ma et~al.(2023)Ma, Zhao, Chen, Wang, Guo, Zhang, Shen, Jiang, and Liu]{ma2023eye}
Chong Ma, Lin Zhao, Yuzhong Chen, Sheng Wang, Lei Guo, Tuo Zhang, Dinggang Shen, Xi~Jiang, and Tianming Liu.
\newblock Eye-gaze-guided vision transformer for rectifying shortcut learning.
\newblock \emph{IEEE Transactions on Medical Imaging}, 2023.

\bibitem[Mani et~al.(2020)Mani, Yoo, Hinthorn, and Russakovsky]{mani2020point}
Arjun Mani, Nobline Yoo, Will Hinthorn, and Olga Russakovsky.
\newblock Point and ask: Incorporating pointing into visual question answering.
\newblock \emph{arXiv preprint arXiv:2011.13681}, 2020.

\bibitem[Niebur \& Koch(1995)Niebur and Koch]{Niebur1995ControlOS}
Ernst Niebur and Christof Koch.
\newblock Control of selective visual attention: Modeling the where pathway.
\newblock In \emph{Neural Information Processing Systems}, 1995.
\newblock URL \url{https://api.semanticscholar.org/CorpusID:7125242}.

\bibitem[OpenAI(2023)]{openai2023gpt4}
OpenAI.
\newblock Gpt-4 technical report, 2023.

\bibitem[Pan et~al.(2016)Pan, Sayrol, i~Nieto, McGuinness, and O’Connor]{Pan2016ShallowAD}
Junting Pan, Elisa Sayrol, Xavier~Giro i~Nieto, Kevin McGuinness, and Noel~E. O’Connor.
\newblock Shallow and deep convolutional networks for saliency prediction.
\newblock \emph{2016 IEEE Conference on Computer Vision and Pattern Recognition (CVPR)}, pp.\  598--606, 2016.
\newblock URL \url{https://api.semanticscholar.org/CorpusID:16408631}.

\bibitem[Peng et~al.(2023)Peng, Wang, Dong, Hao, Huang, Ma, and Wei]{peng2023kosmos}
Zhiliang Peng, Wenhui Wang, Li~Dong, Yaru Hao, Shaohan Huang, Shuming Ma, and Furu Wei.
\newblock Kosmos-2: Grounding multimodal large language models to the world.
\newblock \emph{arXiv preprint arXiv:2306.14824}, 2023.

\bibitem[Piening et~al.(2021)Piening, Piening, Pfeuffer, Esteves, Mittermeier, Prange, Schröder, and Alt]{Piening_2021looking-for-info}
Robin Piening, Robin Piening, Ken Pfeuffer, Augusto Esteves, Tim Mittermeier, Sarah Prange, Philippe Schröder, and Florian Alt.
\newblock Looking for info: Evaluation of gaze based information retrieval in augmented reality.
\newblock \emph{IFIP TC13 International Conference on Human-Computer Interaction}, 2021.
\newblock \doi{10.1007/978-3-030-85623-6_32}.

\bibitem[Pont-Tuset et~al.(2020)Pont-Tuset, Uijlings, Changpinyo, Soricut, and Ferrari]{pont2020connecting}
Jordi Pont-Tuset, Jasper Uijlings, Soravit Changpinyo, Radu Soricut, and Vittorio Ferrari.
\newblock Connecting vision and language with localized narratives.
\newblock In \emph{Computer Vision--ECCV 2020: 16th European Conference, Glasgow, UK, August 23--28, 2020, Proceedings, Part V 16}, pp.\  647--664. Springer, 2020.

\bibitem[Qian et~al.(2023)Qian, Zhang, Song, and Liao]{qian2023gvgnet}
Kun Qian, Zhuoyang Zhang, Wei Song, and Jianfeng Liao.
\newblock Gvgnet: Gaze-directed visual grounding for learning under-specified object referring intention.
\newblock \emph{IEEE Robotics and Automation Letters}, 2023.

\bibitem[Radford et~al.(2021)Radford, Kim, Hallacy, Ramesh, Goh, Agarwal, Sastry, Askell, Mishkin, Clark, Krueger, and Sutskever]{radford2021learning}
Alec Radford, Jong~Wook Kim, Chris Hallacy, Aditya Ramesh, Gabriel Goh, Sandhini Agarwal, Girish Sastry, Amanda Askell, Pamela Mishkin, Jack Clark, Gretchen Krueger, and Ilya Sutskever.
\newblock Learning transferable visual models from natural language supervision, 2021.

\bibitem[Sood et~al.(2021)Sood, Kögel, Strohm, Dhar, and Bulling]{sood21_conll}
Ekta Sood, Fabian Kögel, Florian Strohm, Prajit Dhar, and Andreas Bulling.
\newblock Vqa-mhug: A gaze dataset to study multimodal neural attention in vqa.
\newblock In \emph{Proc. ACL SIGNLL Conference on Computational Natural Language Learning (CoNLL)}, pp.\  27--43. Association for Computational Linguistics, 2021.
\newblock \doi{10.18653/v1/2021.conll-1.3}.

\bibitem[Sugano \& Bulling(2016)Sugano and Bulling]{Sugano2016SeeingWH}
Yusuke Sugano and Andreas Bulling.
\newblock Seeing with humans: Gaze-assisted neural image captioning.
\newblock \emph{ArXiv}, abs/1608.05203, 2016.
\newblock URL \url{https://api.semanticscholar.org/CorpusID:13903325}.

\bibitem[Tanriverdi \& Jacob(2000)Tanriverdi and Jacob]{tanriverdi2000interacting-with-eye}
Vildan Tanriverdi and Robert~JK Jacob.
\newblock Interacting with eye movements in virtual environments.
\newblock In \emph{Proceedings of the SIGCHI conference on Human Factors in Computing Systems}, pp.\  265--272, 2000.

\bibitem[Tavakoli et~al.(2017)Tavakoli, Ahmed, Borji, and Laaksonen]{Tavakoli2017SaliencyRA}
Hamed~Rezazadegan Tavakoli, Fawad Ahmed, Ali Borji, and Jorma~T. Laaksonen.
\newblock Saliency revisited: Analysis of mouse movements versus fixations.
\newblock \emph{2017 IEEE Conference on Computer Vision and Pattern Recognition (CVPR)}, pp.\  6354--6362, 2017.
\newblock URL \url{https://api.semanticscholar.org/CorpusID:12317174}.

\bibitem[Team(2023)]{MosaicML2023Introducing}
MosaicML~NLP Team.
\newblock Introducing mpt-7b: A new standard for open-source, commercially usable llms, 2023.
\newblock URL \url{www.mosaicml.com/blog/mpt-7b}.
\newblock Accessed: 2023-05-05.

\bibitem[Tonsen et~al.(2020)Tonsen, Baumann, and Dierkes]{tonsen2020high}
Marc Tonsen, Chris~Kay Baumann, and Kai Dierkes.
\newblock A high-level description and performance evaluation of pupil invisible.
\newblock \emph{arXiv preprint arXiv:2009.00508}, 2020.

\bibitem[Touvron et~al.(2023)Touvron, Lavril, Izacard, Martinet, Lachaux, Lacroix, Rozi{\`e}re, Goyal, Hambro, Azhar, et~al.]{touvron2023llama}
Hugo Touvron, Thibaut Lavril, Gautier Izacard, Xavier Martinet, Marie-Anne Lachaux, Timoth{\'e}e Lacroix, Baptiste Rozi{\`e}re, Naman Goyal, Eric Hambro, Faisal Azhar, et~al.
\newblock Llama: Open and efficient foundation language models.
\newblock \emph{arXiv preprint arXiv:2302.13971}, 2023.

\bibitem[Vasudevan et~al.(2018)Vasudevan, Dai, and Gool]{Vasudevan2018ObjectRI}
Arun~Balajee Vasudevan, Dengxin Dai, and Luc~Van Gool.
\newblock Object referring in videos with language and human gaze.
\newblock \emph{2018 IEEE/CVF Conference on Computer Vision and Pattern Recognition}, pp.\  4129--4138, 2018.
\newblock URL \url{https://api.semanticscholar.org/CorpusID:4576781}.

\bibitem[Voigtlaender et~al.(2023)Voigtlaender, Changpinyo, Pont-Tuset, Soricut, and Ferrari]{voigtlaender2023connecting}
Paul Voigtlaender, Soravit Changpinyo, Jordi Pont-Tuset, Radu Soricut, and Vittorio Ferrari.
\newblock Connecting vision and language with video localized narratives.
\newblock In \emph{Proceedings of the IEEE/CVF Conference on Computer Vision and Pattern Recognition}, pp.\  2461--2471, 2023.

\bibitem[Wang et~al.(2022)Wang, Yang, Men, Lin, Bai, Li, Ma, Zhou, Zhou, and Yang]{wang2022ofa}
Peng Wang, An~Yang, Rui Men, Junyang Lin, Shuai Bai, Zhikang Li, Jianxin Ma, Chang Zhou, Jingren Zhou, and Hongxia Yang.
\newblock Ofa: Unifying architectures, tasks, and modalities through a simple sequence-to-sequence learning framework.
\newblock In \emph{International Conference on Machine Learning}, pp.\  23318--23340. PMLR, 2022.

\bibitem[Yan et~al.(2021)Yan, Ji, Luo, Zhou, Duan, and Ma]{yan2021control}
Kun Yan, Lei Ji, Huaishao Luo, Ming Zhou, Nan Duan, and Shuai Ma.
\newblock Control image captioning spatially and temporally.
\newblock In \emph{Proceedings of the 59th Annual Meeting of the Association for Computational Linguistics and the 11th International Joint Conference on Natural Language Processing (Volume 1: Long Papers)}, pp.\  2014--2025, 2021.

\bibitem[Yao et~al.(2022)Yao, Chen, Zhang, Ji, Liu, Chua, and Sun]{yao2022pevl}
Yuan Yao, Qianyu Chen, Ao~Zhang, Wei Ji, Zhiyuan Liu, Tat-Seng Chua, and Maosong Sun.
\newblock Pevl: Position-enhanced pre-training and prompt tuning for vision-language models.
\newblock \emph{arXiv preprint arXiv:2205.11169}, 2022.

\bibitem[Yarbus(1967)]{Yarbus1967EyeMA}
Alfred~L. Yarbus.
\newblock Eye movements and vision.
\newblock In \emph{Springer US}, 1967.
\newblock URL \url{https://api.semanticscholar.org/CorpusID:40057049}.

\bibitem[Ye et~al.(2023)Ye, Xu, Xu, Ye, Yan, Zhou, Wang, Hu, Shi, Shi, et~al.]{ye2023mplug}
Qinghao Ye, Haiyang Xu, Guohai Xu, Jiabo Ye, Ming Yan, Yiyang Zhou, Junyang Wang, Anwen Hu, Pengcheng Shi, Yaya Shi, et~al.
\newblock mplug-owl: Modularization empowers large language models with multimodality.
\newblock \emph{arXiv preprint arXiv:2304.14178}, 2023.

\bibitem[Zhang et~al.(2023{\natexlab{a}})Zhang, Han, Zhou, Hu, Yan, Lu, Li, Gao, and Qiao]{zhang2023llama}
Renrui Zhang, Jiaming Han, Aojun Zhou, Xiangfei Hu, Shilin Yan, Pan Lu, Hongsheng Li, Peng Gao, and Yu~Qiao.
\newblock Llama-adapter: Efficient fine-tuning of language models with zero-init attention.
\newblock \emph{arXiv preprint arXiv:2303.16199}, 2023{\natexlab{a}}.

\bibitem[Zhang et~al.(2020)Zhang, Saran, Liu, Zhu, Guo, Niekum, Ballard, and Hayhoe]{Zhang2020HumanGA}
Ruohan Zhang, Akanksha Saran, Bo~Liu, Yifeng Zhu, Sihang Guo, Scott Niekum, Dana~H. Ballard, and Mary~M. Hayhoe.
\newblock Human gaze assisted artificial intelligence: A review.
\newblock \emph{IJCAI : proceedings of the conference}, 2020:\penalty0 4951--4958, 2020.
\newblock URL \url{https://api.semanticscholar.org/CorpusID:220483132}.

\bibitem[Zhang et~al.(2023{\natexlab{b}})Zhang, Sun, Chen, Xiao, Shao, Zhang, Chen, and Luo]{zhang2023gpt4roi}
Shilong Zhang, Peize Sun, Shoufa Chen, Min Xiao, Wenqi Shao, Wenwei Zhang, Kai Chen, and Ping Luo.
\newblock Gpt4roi: Instruction tuning large language model on region-of-interest.
\newblock \emph{arXiv e-prints}, pp.\  arXiv--2307, 2023{\natexlab{b}}.

\bibitem[Zhong et~al.(2022)Zhong, Yang, Zhang, Li, Codella, Li, Zhou, Dai, Yuan, Li, et~al.]{zhong2022regionclip}
Yiwu Zhong, Jianwei Yang, Pengchuan Zhang, Chunyuan Li, Noel Codella, Liunian~Harold Li, Luowei Zhou, Xiyang Dai, Lu~Yuan, Yin Li, et~al.
\newblock Regionclip: Region-based language-image pretraining.
\newblock In \emph{Proceedings of the IEEE/CVF Conference on Computer Vision and Pattern Recognition}, pp.\  16793--16803, 2022.

\bibitem[Zhou et~al.(2023)Zhou, Yu, Zhang, Wu, Wang, and Wang]{zhou2023regionblip}
Qiang Zhou, Chaohui Yu, Shaofeng Zhang, Sitong Wu, Zhibing Wang, and Fan Wang.
\newblock Regionblip: A unified multi-modal pre-training framework for holistic and regional comprehension.
\newblock \emph{arXiv preprint arXiv:2308.02299}, 2023.

\bibitem[Zhu et~al.(2023)Zhu, Chen, Shen, Li, and Elhoseiny]{zhu2023minigpt}
Deyao Zhu, Jun Chen, Xiaoqian Shen, Xiang Li, and Mohamed Elhoseiny.
\newblock Minigpt-4: Enhancing vision-language understanding with advanced large language models.
\newblock \emph{arXiv preprint arXiv:2304.10592}, 2023.

\bibitem[Zhu et~al.(2016)Zhu, Groth, Bernstein, and Fei-Fei]{zhu2016visual7w}
Yuke Zhu, Oliver Groth, Michael Bernstein, and Li~Fei-Fei.
\newblock Visual7w: Grounded question answering in images.
\newblock In \emph{Proceedings of the IEEE conference on computer vision and pattern recognition}, pp.\  4995--5004, 2016.

\end{thebibliography}
\bibliographystyle{iclr2024_conference}

\appendix
\section{Limitation and Discussion}
\subsection{Hallucination}

During inference, our method occasionally exhibits hallucinations related to image content. This issue may stem from the limited number of training samples and the imperfect integration of visual hidden distributions into the language decoding process. Recognizing the potential for further scaling of our method, we consider addressing these hallucinations as a future research direction.
\subsection{Comparison to GPT-4-V}

With the release of GPT-4-V, which features vision capabilities, during the submission of this paper, we find it pertinent to include a brief discussion on the relationship between our work and GPT-4-V. Although GPT-4-V demonstrates remarkable visual capabilities in its demos, it surpasses our method in terms of visual understanding abilities. Nonetheless, our work remains valuable as it presents an effective approach to incorporating user sensory information for generating more relevant responses.

As reported, GPT-4-V occasionally struggles to accurately capture a user's intent when referencing specific elements within an image, prompting the design of an interface that allows users to directly draw sketches for highlighting purposes. Our method has the potential to enhance this user experience and can be extended to more dynamic scenarios such as virtual reality and augmented reality.
\begin{figure}[t]
\centering  {\includegraphics[width=1.0\linewidth]{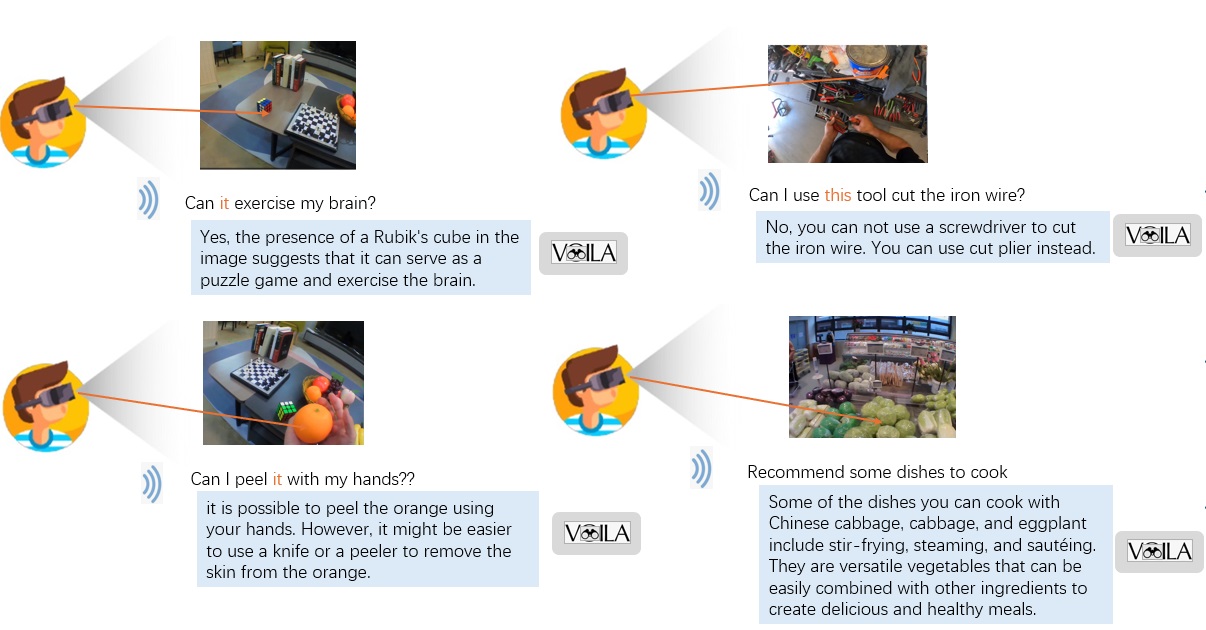}}    \caption{Different Scenarios in the Future
   }
   \label{fig:voila}
\end{figure}

\begin{figure}

    \centering
    \begin{AIbox}{Qualitative Case Study}
        \scriptsize
        \begin{tabular}{*{6}{p{0.14\linewidth}}}
            \multicolumn{1}{p{0.14\linewidth}}{\bfseries Original} &
            \multicolumn{1}{p{0.14\linewidth}}{\bfseries Gaze Heatmap} &
            \multicolumn{1}{p{0.14\linewidth}}{\bfseries Ground Truth} &
            \multicolumn{1}{p{0.14\linewidth}}{\bfseries Otter} &
            \multicolumn{1}{p{0.14\linewidth}}{\bfseries Kosmos-2} &
            \multicolumn{1}{p{0.14\linewidth}}{\bfseries Voila} \\
            \parbox[c]{\linewidth}{\includegraphics[width=\linewidth]{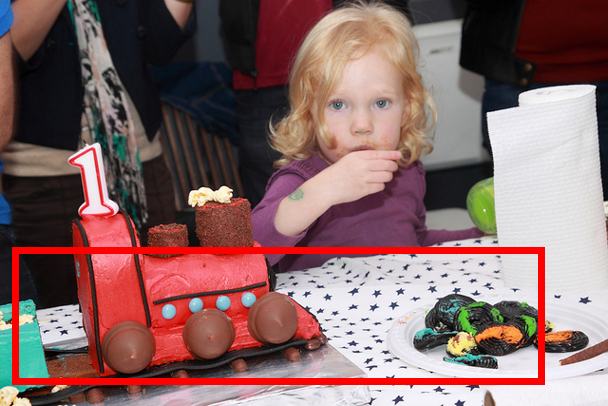}} &
            \parbox[c]{\linewidth}{\includegraphics[width=\linewidth]{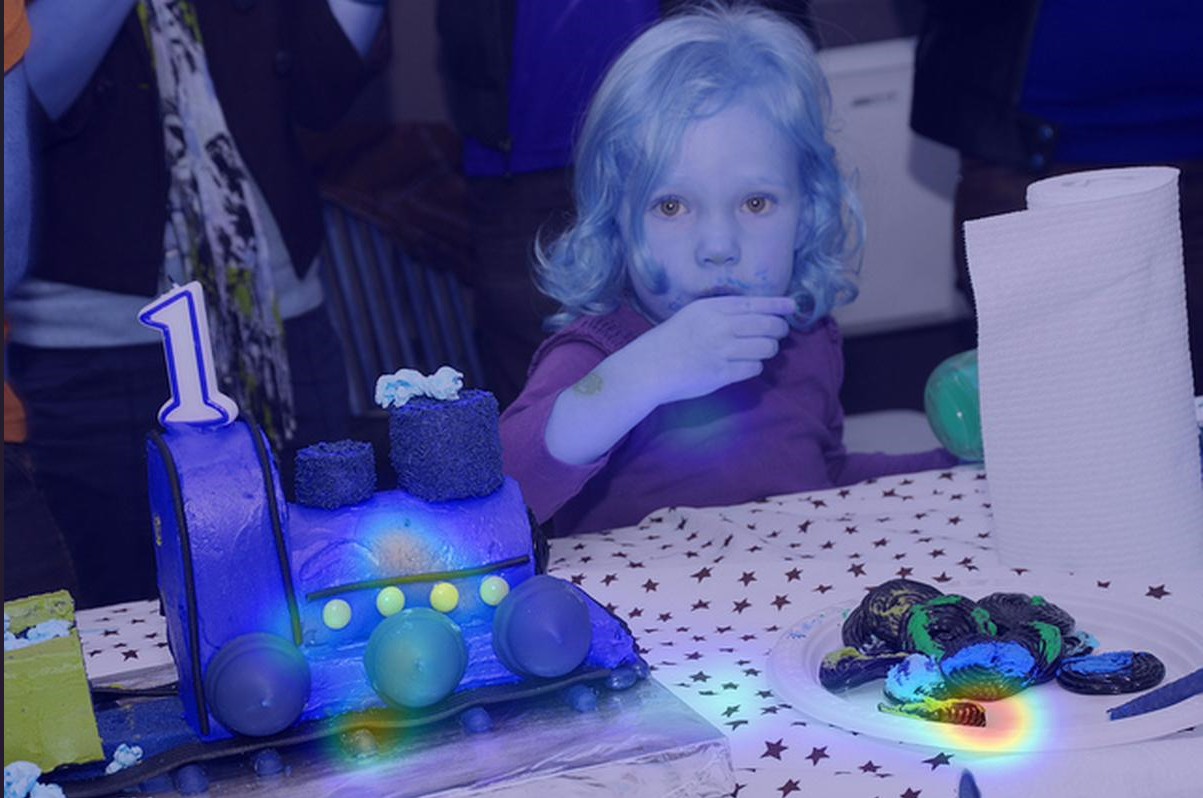}} &
            \parbox[c]{\linewidth}{{\bf Query}: What colors are the cakes?\\
            {\bf Answer}: The cakes on the table are in red, brown, and orange colors.}&
            \parbox[c]{\linewidth}{There are two cakes in the image. The first cake is white, and the second cake has a red color.} &
            \parbox[c]{\linewidth}{The cakes are decorated with red and black colors } &
            \parbox[c]{\linewidth}{The cakes on the table are in red, brown, and orange colors.
} \\
            \parbox[c]{\linewidth}{\includegraphics[width=\linewidth]{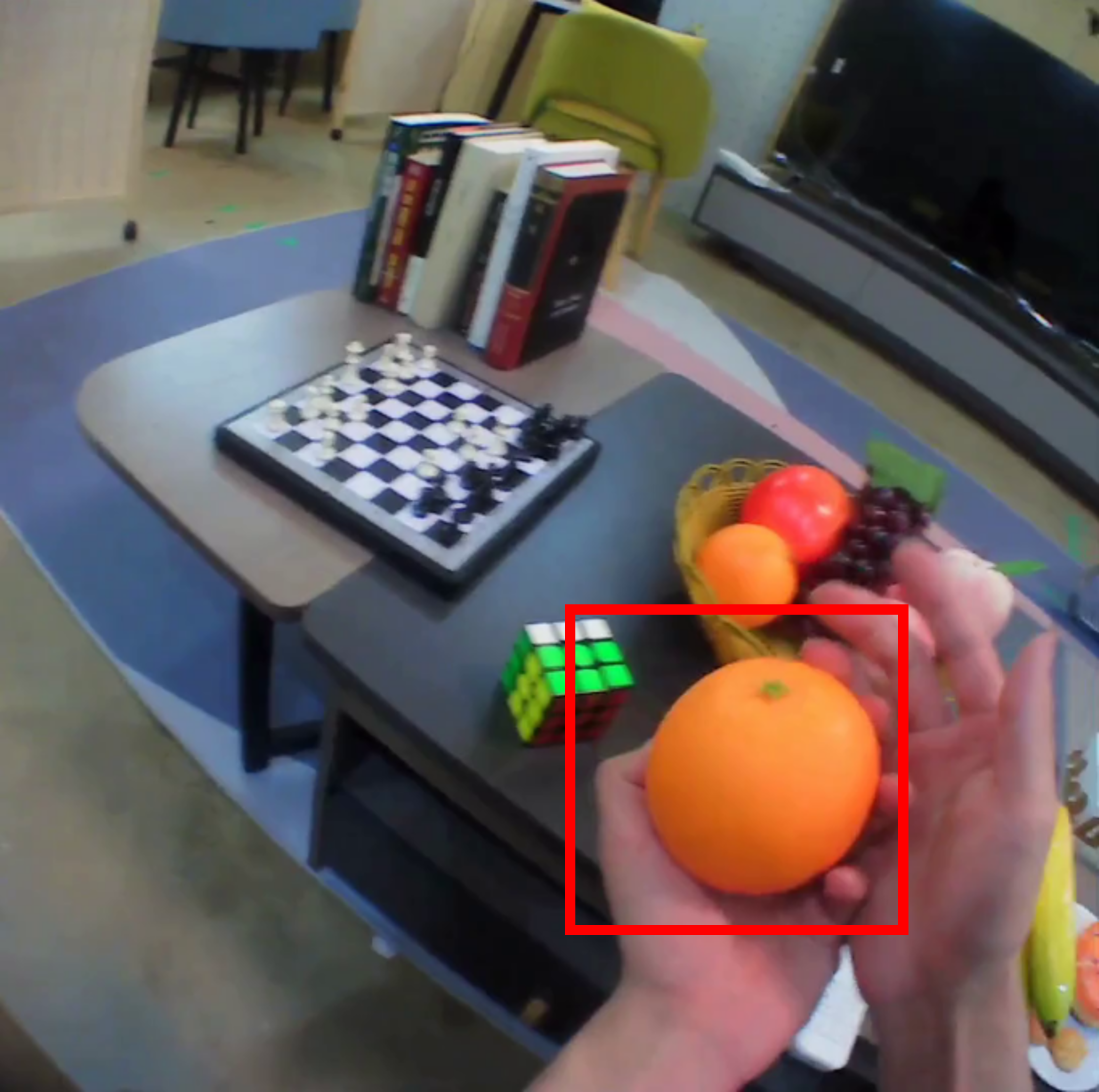}} &
            \parbox[c]{\linewidth}{\includegraphics[width=\linewidth]{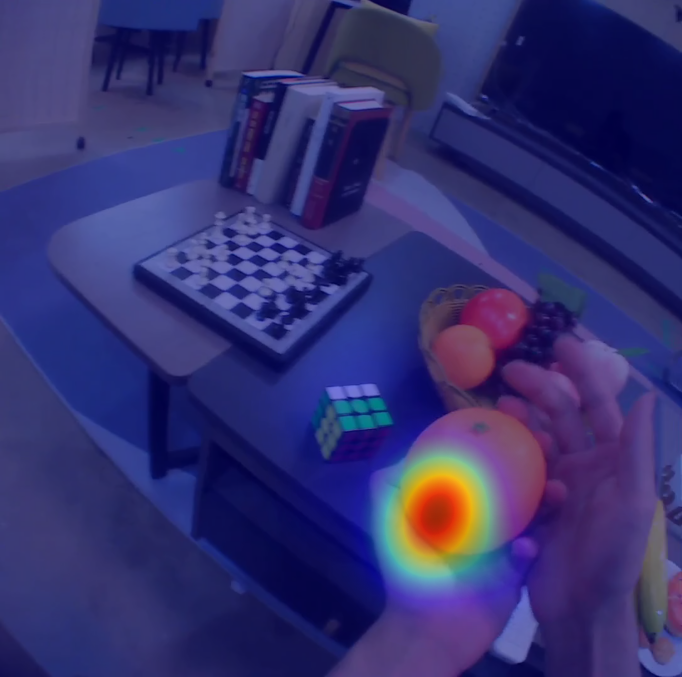}} &
            \parbox[c]{\linewidth}{{\bf Question:} Can I peel it with my hands?
\textbf{Answer:} Yes, you can peel an orange with your hands...} &
            \parbox[c]{\linewidth}{I'm sorry, but I couldn't understand the answer. Can you please rephrase or provide more context?.} &
            \parbox[c]{\linewidth}{Yes , you can peel an orange with your hands 
} &
            \parbox[c]{\linewidth}{Yes, it is possible to peel the orange using your hands. However, it might be easier to use a knife or a peeler to remove the skin from the orange.} \\
            \parbox[c]{\linewidth}{\includegraphics[width=\linewidth]{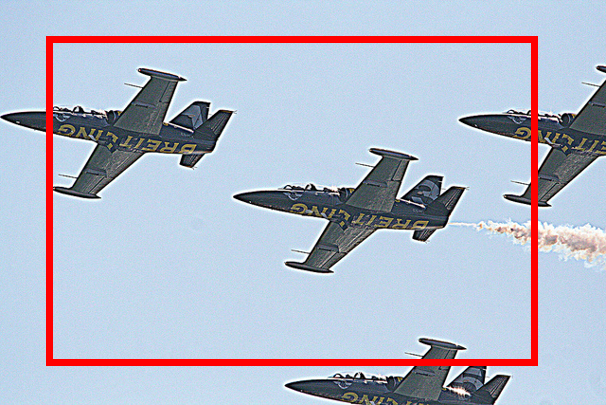}} &
            \parbox[c]{\linewidth}{\includegraphics[width=\linewidth]{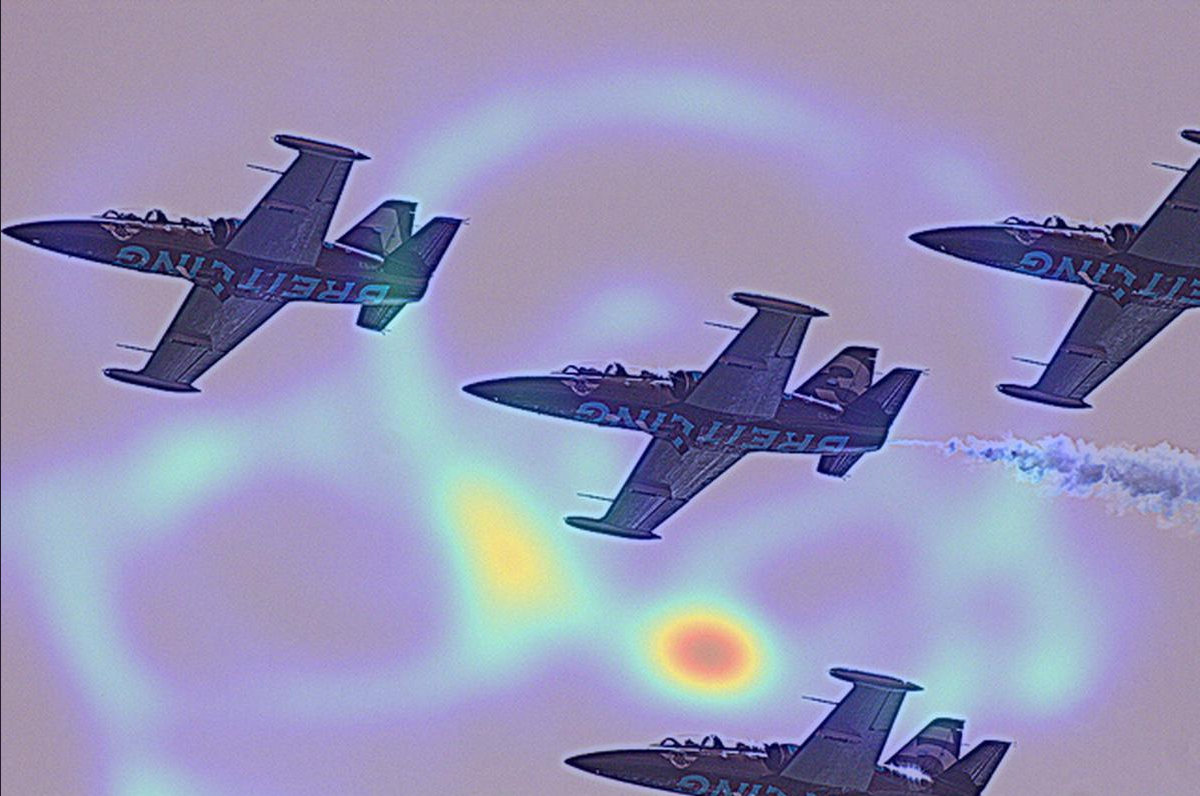}} &
            \parbox[c]{\linewidth}{\textbf{Question:} What's the color?
\textbf{Answer:} The color of the sky in the background is blue
} &
            \parbox[c]{\linewidth}{The color of the plane is white.
} &
            \parbox[c]{\linewidth}{The color of the planes in the image is blue
} &
            \parbox[c]{\linewidth}{The background is blue in color.
} \\
            \parbox[c]{\linewidth}{\includegraphics[width=\linewidth]{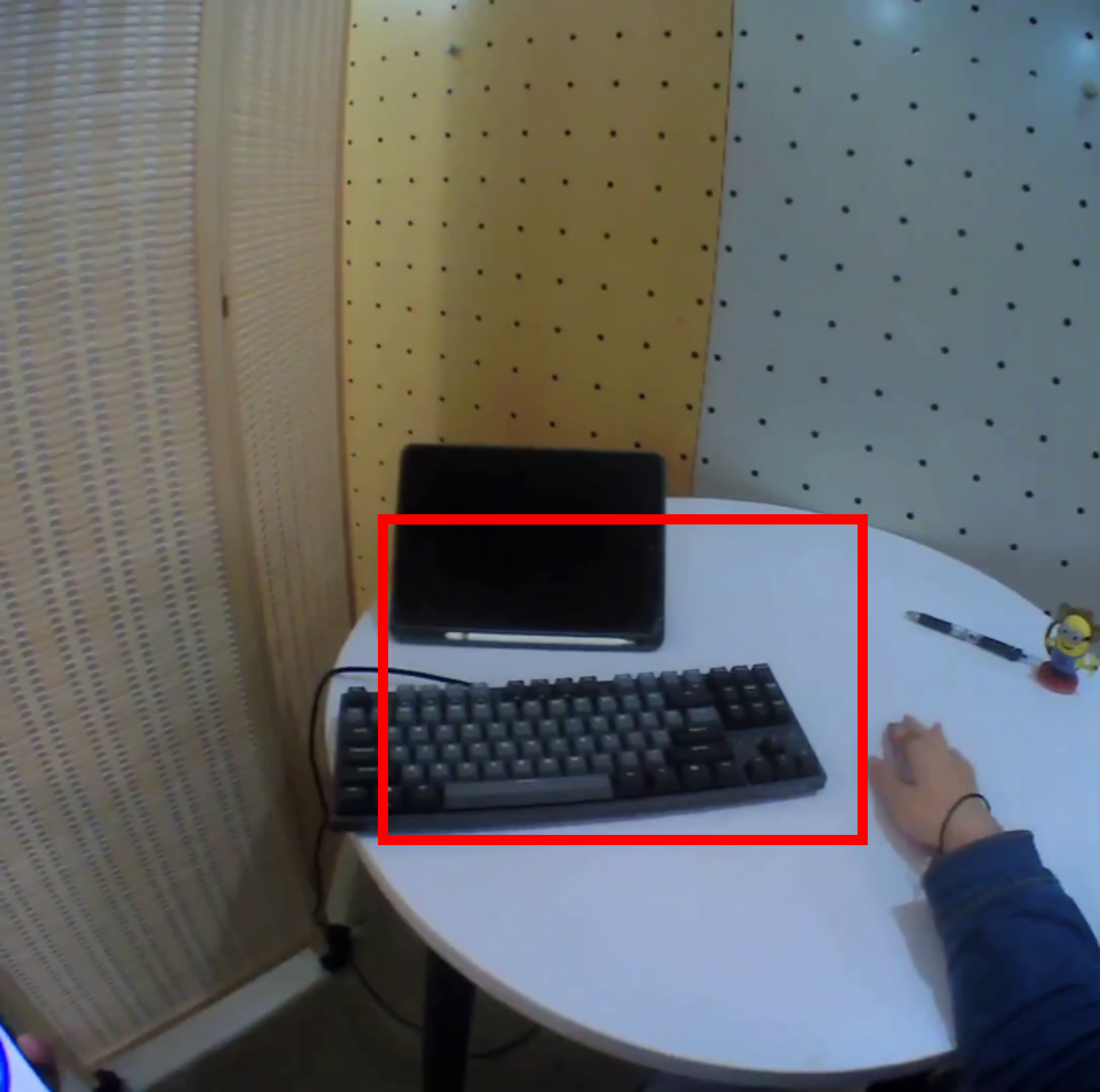}} &
            \parbox[c]{\linewidth}{\includegraphics[width=\linewidth]{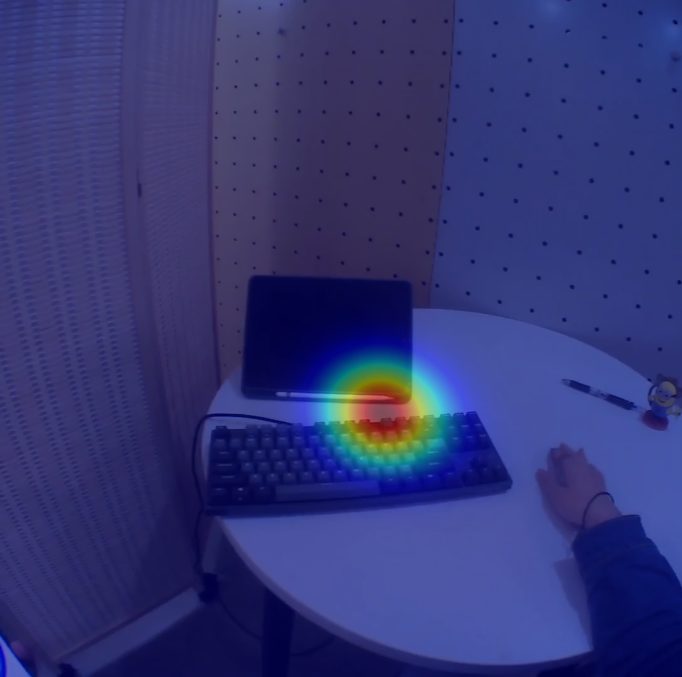}} &
            \parbox[c]{\linewidth}{\textbf{Question:} How to reduce eye fatigue?
\textbf{Answer:} To reduce eye fatigue while using a laptop, take regular breaks, follow the 20-20-20 rule … adjust screen brightness and contrast, …
} &
            \parbox[c]{\linewidth}{The most common cause of eye fatigue is overexposure to bright light …} &
            \parbox[c]{\linewidth}{Keyboard} &
            \parbox[c]{\linewidth}{To minimize eye fatigue while working on a laptop, consider the following steps: 1. Adjust the screen‘s brightness to a comfortable level …} \\
            \parbox[c]{\linewidth}{\includegraphics[width=\linewidth]{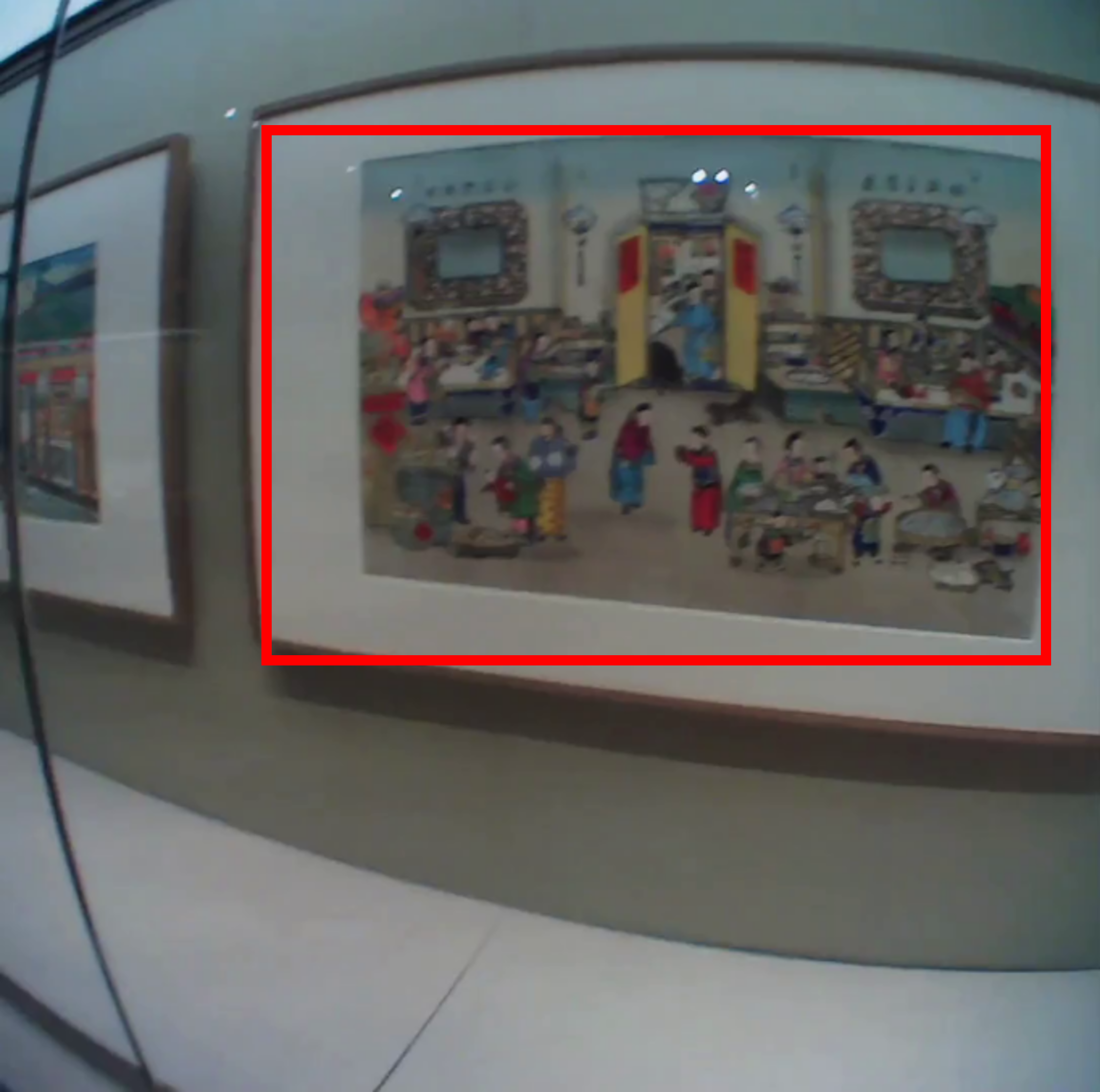}} &
            \parbox[c]{\linewidth}{\includegraphics[width=\linewidth]{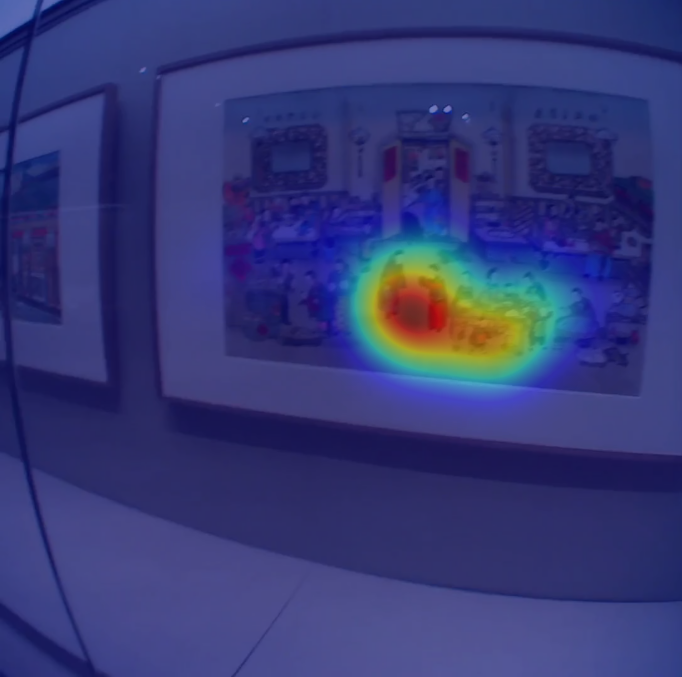}} &
            \parbox[c]{\linewidth}{\textbf{Question:} How many figures are there in this painting?
\textbf{Answer:} According to what I see, the painting seems to contain 23 figures.
} &
            \parbox[c]{\linewidth}{There are two figures in this image.
} &
            \parbox[c]{\linewidth}{There are two figures in the painting .
} &
            \parbox[c]{\linewidth}{There are two people standing in the painting.
} \\
        \end{tabular}
    \end{AIbox}
    \caption{Qualitative Case Study: Top: We show successful predictions of all models. Middle: We demonstrate the problems of baseline models compared with VOILA including coreference queries, gaze grounding methods, etc. Bottom: We display hard challenges for all models.
}
\label{fig:casestudy}
\end{figure}

\section{Gaze Data Collection}
\label{datacollect}
In order to examine the gaze patterns of users, we conducted a preliminary study where participants engaged in one of three daily situations: grocery shopping, visiting a museum, or engaging in domestic activities. Participants were instructed to perform queries specific to each scenario, as if the system could provide the desired information. Our research involved 21 participants (8, 5, and 8 for each respective scenario), consisting of 13 males and 8 females, ranging in age from 19 to 30 years old (with a standard deviation of 3.18). The study took each participant between 90 and 150 minutes to finish, and they were compensated at a rate of 15 USD per hour for their involvement. As a result, we get 548 minutes of gaze recording.
{The Pupil Labs Invisible~\cite{tonsen2020high} is a gaze-tracking smart glasses system that has been widely used for research purposes. It is equipped with gaze sensors, an egocentric camera, a microphone, and an inertial measurement unit (which was not used in this work). Participants were asked to wear the Pupil Labs Invisible glasses without any headwear that could obstruct the sensors on the glasses. Since the Invisible glasses require a connection to a mobile phone for operation, we instructed participants to keep the phone in their pockets to minimize potential distractions. Data was continuously recorded as participants engaged in their chosen scenario.}
\begin{table}[]
    \centering
    \begin{tabular}{c|c}
    \toprule
        \multicolumn{1}{c}{\textbf{Supermarket shopping}} & \multicolumn{1}{c}{\textbf{Domestic living}} \\
         Task  & Task \\
    \midrule
    Comparison  & Appliance Malfunction  \\
    Completing Recipe  & Activity \& Health \\
    Recommend  & Snack \& Fruits  \\
    Knowledge & Dressing Advice  \\
    Decision Making  & Entertainment\\
    Strengthen Decision  & Small Talk  \\
    \bottomrule
    \end{tabular}
    \caption{{
Guiding for User in VOILA-GAZE Collection, note this guide aims to facilitate and inspire users to generate questions related to data collection, rather than imposing strict limitations on the scope of their inquiries.}}
    \label{tab:task_descriptions}
\end{table}
\begin{figure}
    \centering
    \includegraphics[width=\linewidth]{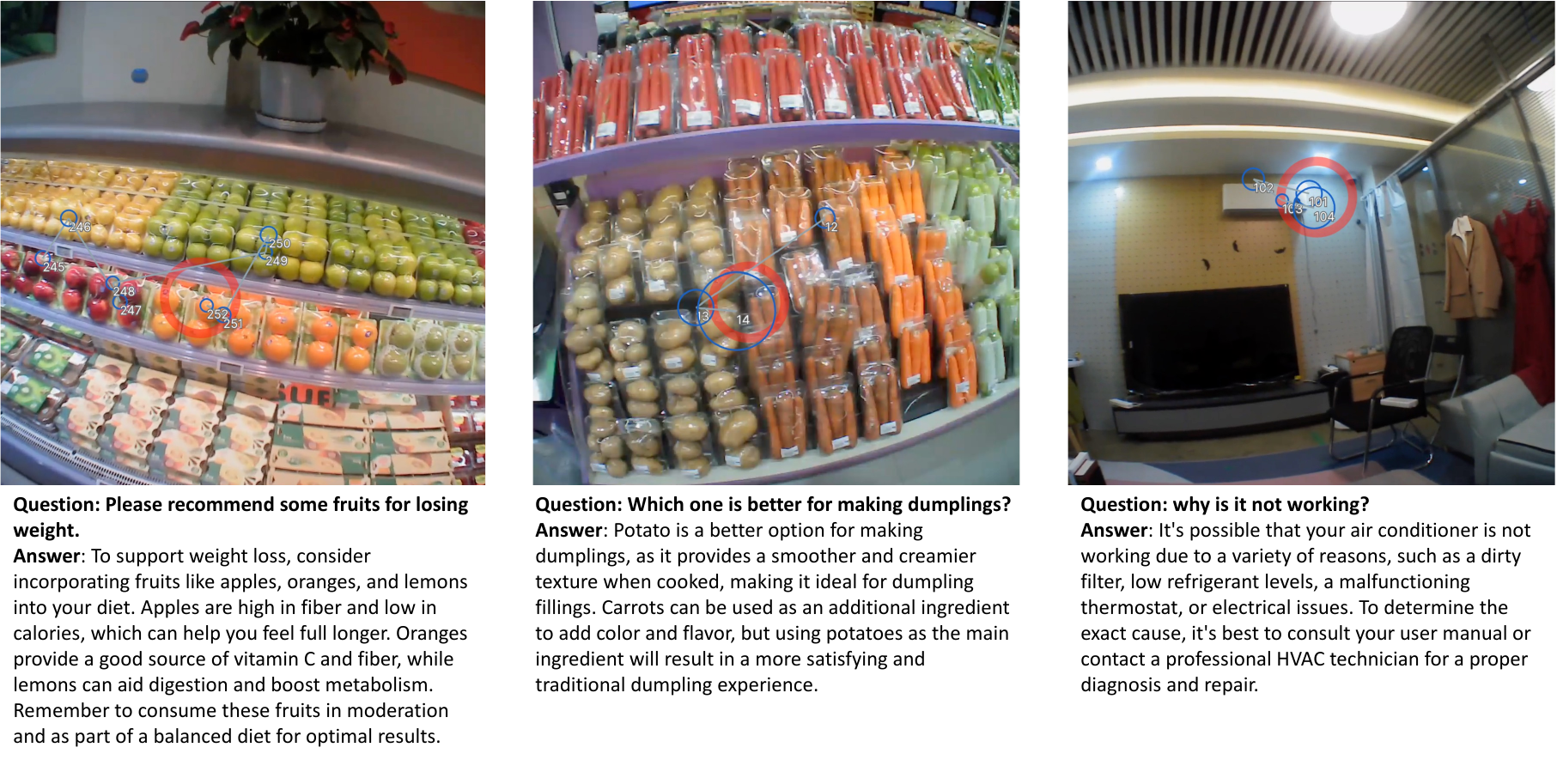}
    \caption{Data sample from VOILA-GAZE}
    \label{fig:voila-gaze-example}
\end{figure}
\section{Details for Automatic Data Annotation}
\label{Prompt}
Figure~\ref{fig:prompt} shows the system prompt and in-context examples for Automatic Data Annotation
 \begin{figure}

    \begin{AIbox}{Prompt for Automatic Referring Annotation}

    {\small \bf System Prompt:} {\scriptsize As an AI visual assistant, your task is to analyze a single view as if you are directly observing it. You will be provided with background and referable sentences describing the view, but your responses should be based on your observations rather than the descriptions.
    First, annotate the referable sentence using the \(<\)Q\#\(>\)\(<\)/Q\#\(>\) tag, where \# is the tag number, to emphasize the section you want to discuss. Note: DO NOT annotate background sentences. Then, engage in a conversation with a person inquiring about the view, focusing on the tagged elements while considering the background information. Use the \(<\)Q\#\(>\) tag for questions related to specific tags in the sentence, and the \(<\)Q\(>\) tag for other questions. The conversation should resemble the tone of a visual AI assistant examining the view and answering questions. Include a variety of questions and their corresponding answers.
    Your questions should cover various aspects of the view's visual content, such as object types, object count, actions, locations, and relative positions between objects. Ensure your questions have definite answers: (1) the view clearly displays the content in question, allowing for a confident response; or (2) it can be confidently determined from the view that the content is not present. Avoid questions that cannot be answered with certainty.
    Try your best to incorporate complex questions relevant to the view's content, including background information about the objects or discussions about events occurring within the view. Refrain from asking about uncertain details. When using \(<\)Q\#\(>\), follow the original question with an indirect question, as if the questioner is looking at the area of interest; use pronouns to refer to the relevant part, and omit some context in the original question if it can be inferred from the user's eye gaze fixation. When answering complex questions, provide detailed responses, including examples or reasoning steps, to create a convincing and well-organized explanation. Use multiple paragraphs if necessary. Make the answer comprehensive and informative. Throughout the conversation, maintain the impression that you are directly observing the view rather than reading descriptions. Do not mention any image or picture or text in the question and answer, as if you are equipped in a HMD and sharing the same view with the user.} 
    \tcbline
    {\small \bf In context Example:} \\
    {\scriptsize 
    \parbox[t]{0.15\linewidth}{
        {\bf Role: user}\\
        Background: There is a movie theater that displays the show times above the doors.A hotel for dogs in the snow in winter. \\Referable:A red fire hydrant is deep in the snow.The fire hydrant is in the snow near a recently plowed sidewalk.This city has had a very hard winter with snow.\\
        }
        \hspace{0.2in}
        \parbox[t]{0.8\linewidth}{
        {
            {\bf Role: assistant} \\
            Refer:\(<\)Q1\(>\)A red fire hydrant\(<\)/Q1\(>\) is \(<\)Q2\(>\)deep in the snow\(<\)/Q2\(>\).\(<\)Q3\(>\)The fire hydrant is in the snow near a recently plowed sidewalk.\(<\)/Q3\(>\)\(<\)Q4\(>\)This city has had a very hard winter with snow.\(<\)/Q4\(>\)\\===\\Question:\\\(<\)Q1\(>\)What is the red thing?\\===\\Indirect Question:\\What is it?\\===\\Answer:\\It is a red fire hydrant.\\===\\Question:\\ \(<\)Q2\(>\)Is the fire hydrant fully covered by snow or partially visible?\\===\\Indirect Question:\\Is it fully covered by snow or partially visible?\\===\\Answer:\\The fire hydrant is partially visible, with a significant portion of it covered by the deep snow.\\===\\Question:\\\(<\)Q3\(>\)What is the condition of the sidewalk near the fire hydrant?\\===\\Indirect Question:\\How's the condition of the sidewalk there?\\===\\Answer:\\The sidewalk near the fire hydrant has been recently plowed, so it is relatively clear of snow compared to the surrounding area.\\===\\Question:\\\(<\)Q4\(>\)What does this scene suggests about the winter season in the city?\\===\\Indirect Question:\\How's the winter here?\\===\\Answer:\\It suggests that the winter season in the city has been quite harsh and challenging. The significant snow accumulation seen in the view indicates that the city has experienced heavy snowfall, which can lead to various difficulties and disruptions for its residents. \\For instance, the deep snow covering the fire hydrant could potentially hinder its accessibility in case of emergencies, which poses safety concerns. The recently plowed sidewalk implies that the city's maintenance crews have been actively working to keep the walkways clear and safe for pedestrians, but the sheer amount of snowfall might make it difficult to keep up with the continuous snow removal efforts. \\Furthermore, such extreme winter conditions can impact transportation, leading to traffic issues, delays in public transit, and increased risks of accidents due to slippery roads. It can also cause problems for businesses and schools, as people might face challenges commuting to work or attending classes. \\In conclusion, the red fire hydrant deep in the snow and the recently plowed sidewalk suggest that the city has faced a particularly severe winter season, with substantial snowfall that has likely caused various challenges and disruptions for its residents and infrastructure."
        }}
        }
    \end{AIbox}
    \caption{Prompt for Automatic Referring Annotation}
 \label{fig:prompt}
    \end{figure}

\begin{figure}
    \centering
    \includegraphics[width=\linewidth]{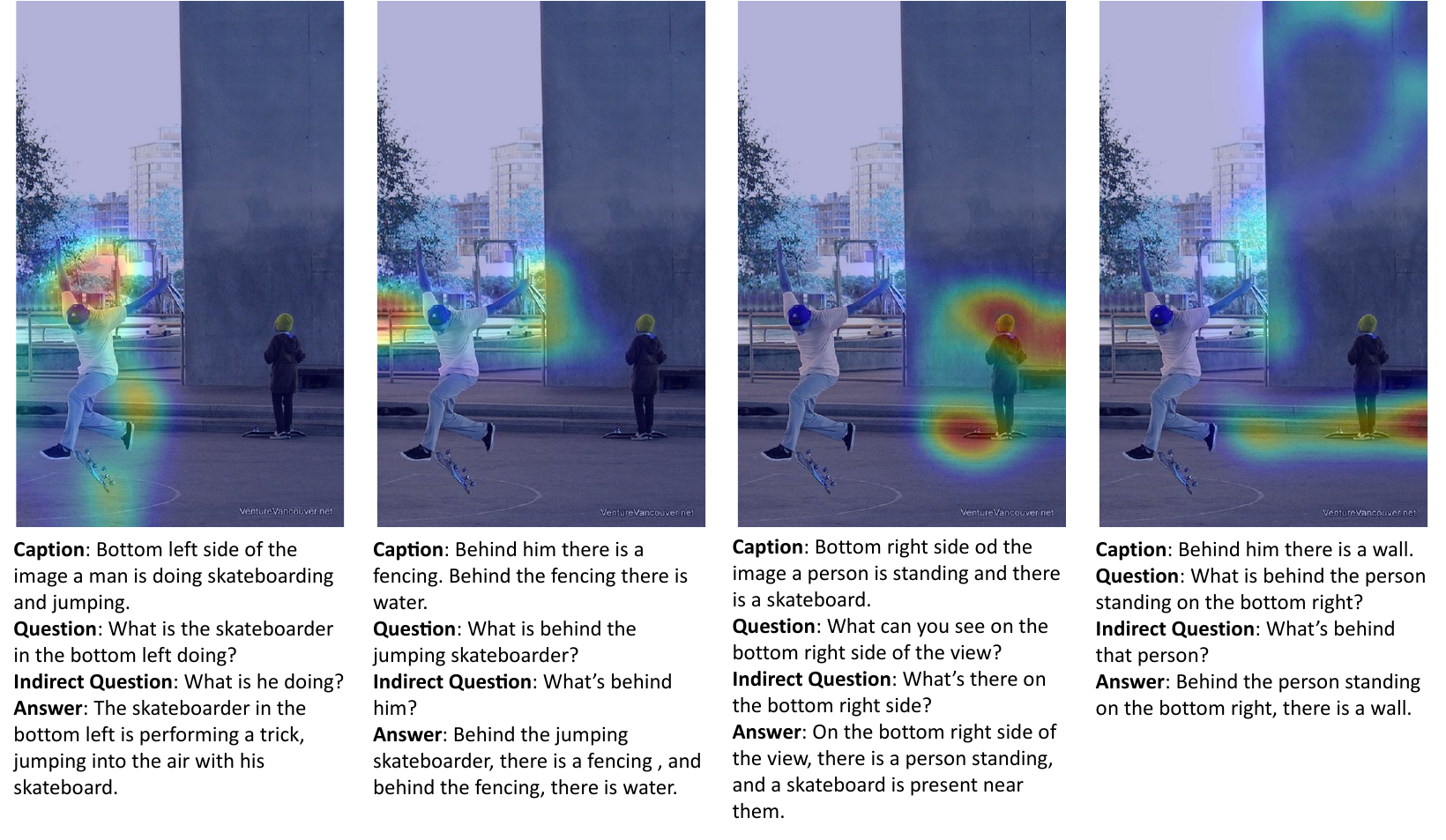}\\
    \includegraphics[width=\linewidth]{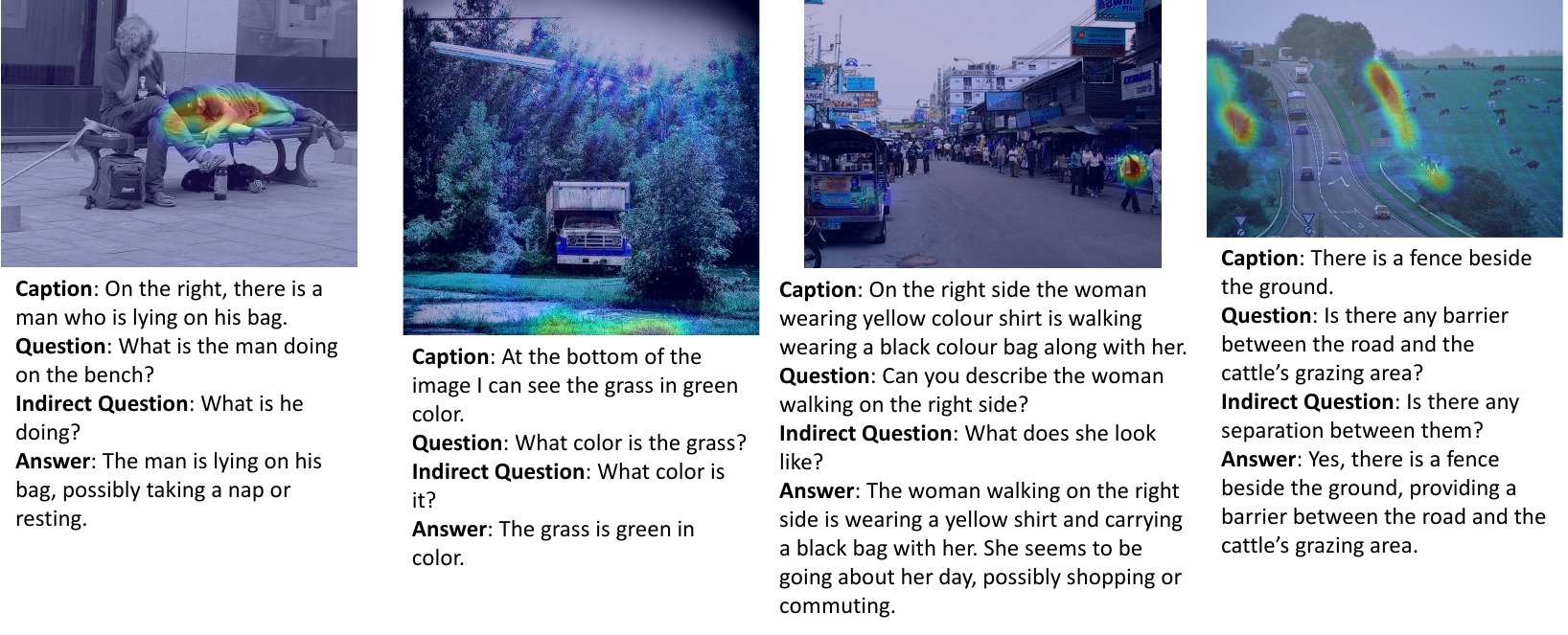}
    \caption{Annotated Example of VOILA-COCO}
    \label{fig:voila-coco-example}
\end{figure}
\section{Implementation Details }
In this section, we describe the implementation details of our model. The model architecture is derived from Otter, which combines a text model and a vision model. The text model is an instance of MPTForCausalLM 7B and the vision model is based on the CLIP ViT-Large model with a patch size of 14.

\begin{itemize}
    \item \textbf{Text Model Configuration:} The text model has 32 layers, each with 32 attention heads. The model has a hidden size and $d_{model}$ of 4096. The attention mechanism uses multi-head attention with torch implementation and alibi attention. The model uses learned position embeddings and low-precision layer normalization. The model is trained with a maximum sequence length of 2048 and a vocabulary size of 50432. The tokenizer used is EleutherAI/gpt-neox-20b. The model's torch data type is set to bfloat16.
    
    \item \textbf{Vision Model Configuration:} The vision model has 24 hidden layers and a hidden size of 1024. It uses 16 attention heads and an intermediate size of 4096. The activation function is quick\_gelu. The model uses an image size of 224 and 3 input channels. The patch size is 14, and the projection dimension is 512. The layer normalization epsilon is set to $1\times10^{-5}$.
    
    \item \textbf{Voila Configuration:} The architecture uses cross attention every 4 layers and only attends to previous layers. The model's torch data type is set to float32. Media placement augmentation is enabled during training.
\end{itemize}

For initialization, we use the Kaiming normal method with fan-in mode, ReLU nonlinearity, and a standard deviation of 0.02. 

For optimization, we employ the AdamW optimizer\cite{kingma2017adam} with a starting learning rate of 1e-5 and a batch size of 4. We train Voila for three epochs, scheduling the learning rate using a cosine annealing scheduler. To prevent exploding gradients, we apply gradient clipping with a threshold of 1.0.
\section{GPT-RANK}
Figure~\ref{GPT-RANK} presents our prompt and evaluation procedure for GPT-RANKING
 \begin{figure}

    \begin{AIbox}{GPT Ranking}

    {\small \bf Overall System Prompt:} {\scriptsize Given a question along with the ground truth description and answer of an image, evaluate the two provided candidate answers. Determine which answer is factually accurate, logical, and helpful to the user. if you think anwser 1 is better, respond with -1, if answer 2 is better respond with 1, if you think the result is tie, output 0. Only respond with either '-1' or '0' or '1' to indicate your choice.} 
    \tcbline
        {\small \bf Helpfulness System Prompt:} {\scriptsize Given a question along with the ground truth description and answer of an image, evaluate the two provided candidate answers. Determine which answer is actually solves the user problem and more helpful to the user. if you think anwser 1 is better, respond with -1, if answer 2 is better respond with 1, if you think the result is tie, output 0. Only respond with either '-1' or '0' or '1' to indicate your choice.} 
    \tcbline
        {\small \bf Fact Grounding System Prompt:} {\scriptsize Given a question along with the ground truth description and answer of an image, evaluate the two provided candidate answers. Determine which answer is factually grounded to the Fact provided. if you think anwser 1 is better, respond with -1, if answer 2 is better respond with 1, if you think the result is tie, output 0. Only respond with either '-1' or '0' or '1' to indicate your choice.} 
    \tcbline

    \begin{algorithmic}[1]
    \For{\textbf{each} key \textbf{in} keys}

        \State answer1 $\gets$ model1[key]["response"]
        \State answer2 $\gets$ model2[key]["response"]
        \State fact, gt\_answer, question $\gets$ dataset[key]
        \State prompt1 $\gets$ CreatePrompt(question, fact, gt\_answer, answer1, answer2)
        \State response\_1 $\gets$ CallGPT(prompt1)
        \State prompt2 $\gets$ CreatePrompt(question, fact, gt\_answer, answer2, answer1)
        \State response\_2 $\gets$ CallGPT(prompt2)
        \State score $\gets$ ComputeScore(response\_1, -1 * response\_2)
        \State Append scores with score
    \EndFor
    \end{algorithmic}

    \end{AIbox}
    \caption{GPT-RANKING Procedure}
\label{GPT-RANK}
    \end{figure}

\section{Ablations on how to incorporate gaze}
\begin{figure}
    \centering
    \includegraphics[width=0.45\linewidth]{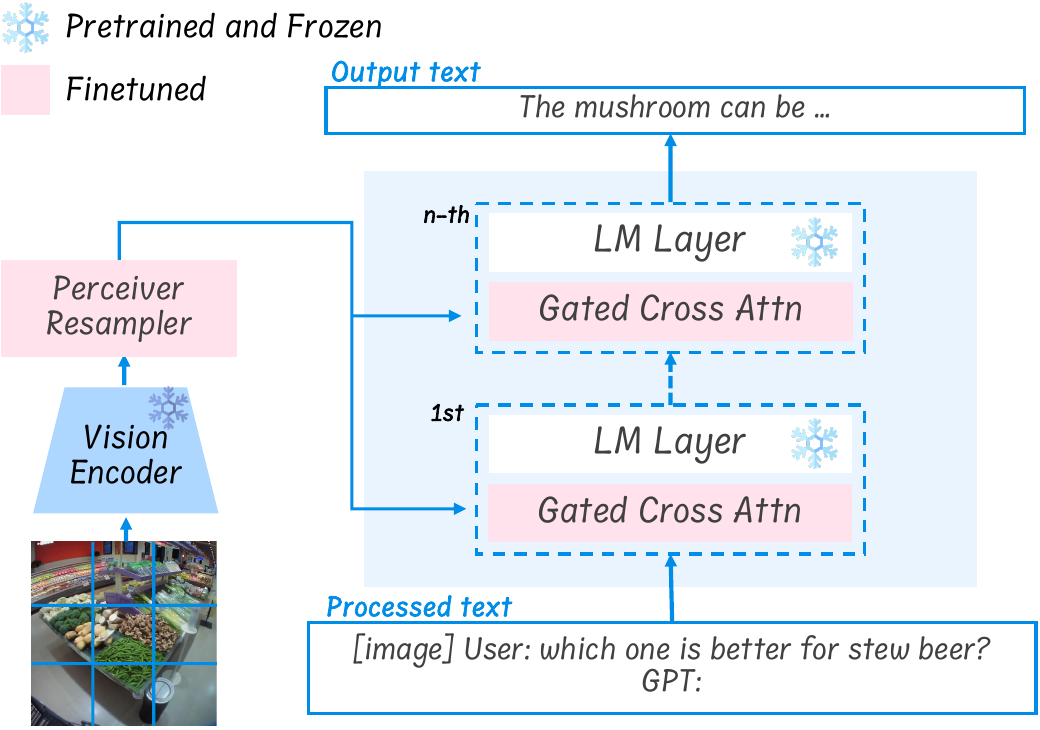} \includegraphics[width=0.45\linewidth]{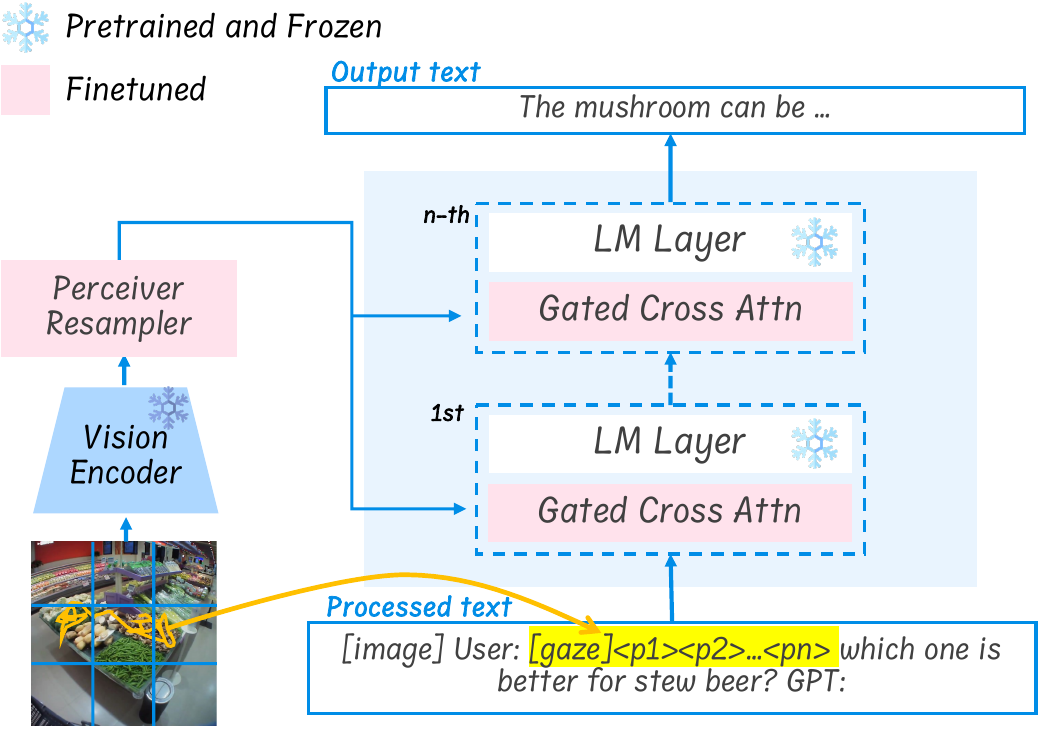}\\
\includegraphics[width=0.45\linewidth]{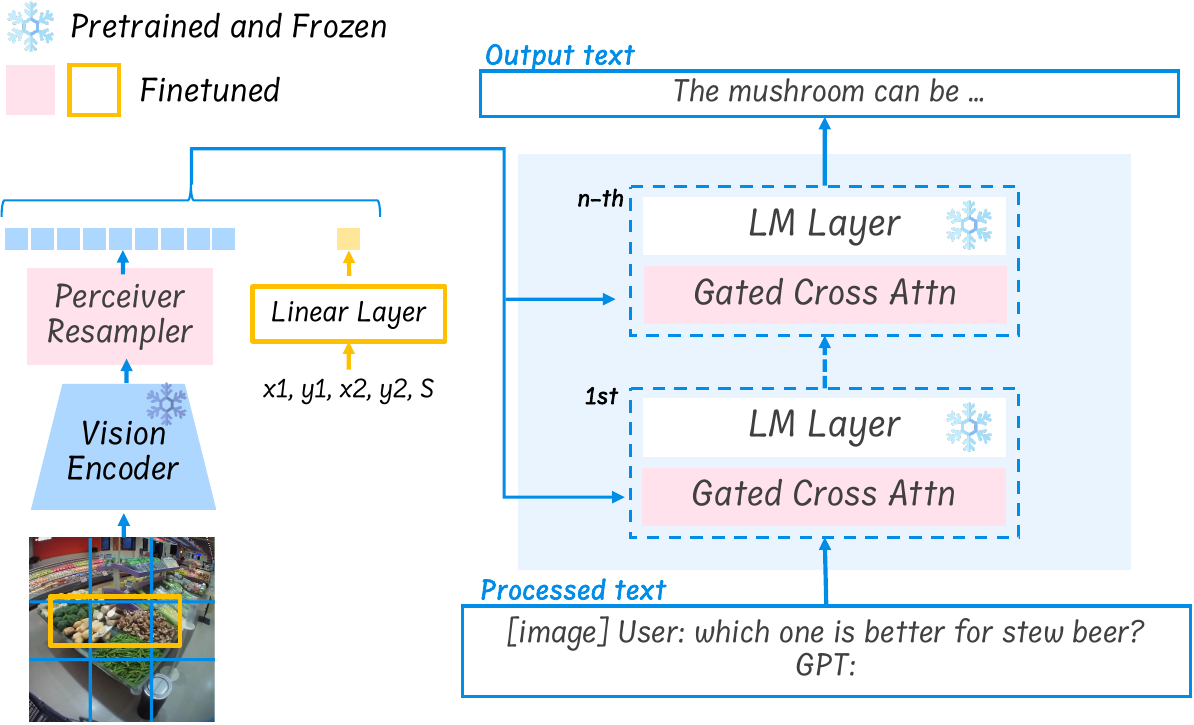}\includegraphics[width=0.45\linewidth]{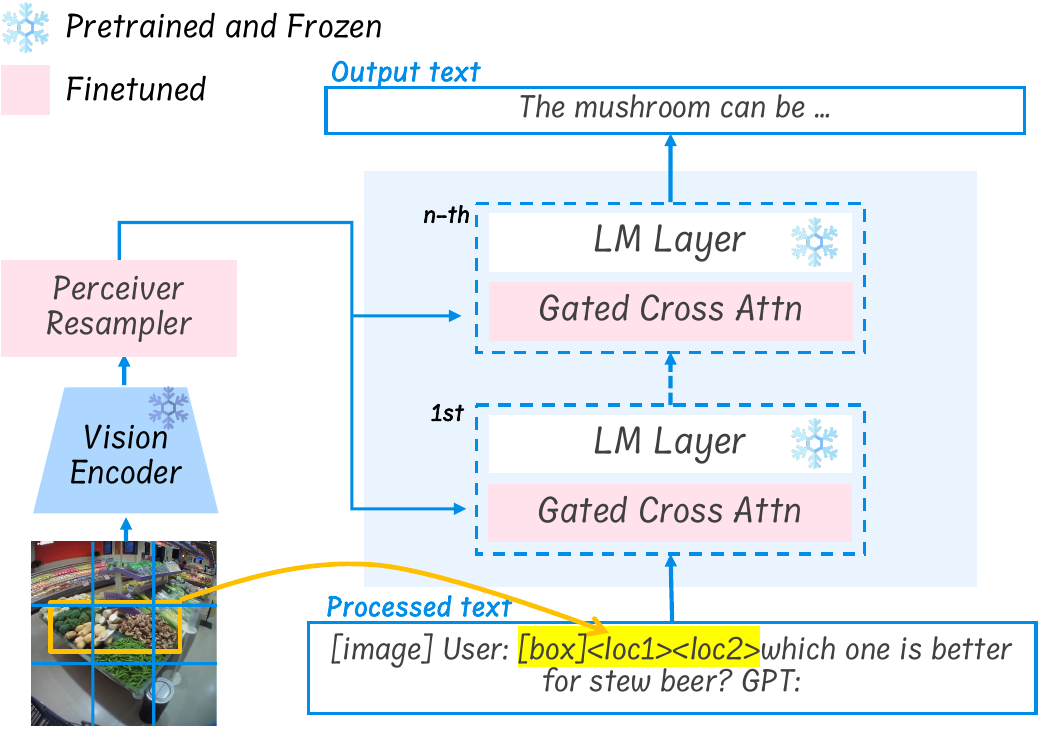} \\
    \caption{Different model design on how to incorporate gaze.Top-left: Otter-base, Top-right: Gaze as discrete position tokens.
Bottom-left: Gaze \emph{bounding box} as image patch 
Bottom-right:Gaze \emph{bounding box} as discrete position tokens }
    \label{fig:model-design}
\end{figure}

Figure \ref{fig:model-design} shows different approaches exist for incorporating gaze data.
\section{Related Work}

\subsection{Multimodal Large Language Model}

Recent research works for multimodal vision and language tasks rely on multimodal large language models(MLLM) and demonstrated superior performance. One line is to learn \textbf{alignment} between the vision input to text token for LLM. LLaVA \cite{liu2023visual} directly feed visual features to the LLM using only a learnable fully connected layer. BLIP-2 \cite{li2023blip} proposed Q-Former to extract a fixed number of query features from visual features, which are aligned with the text in frozen LLM. Another direction is to design \textbf{learnable interaction layers} to attend the vision input to the frozen LLM layers. 
Flamingo\cite{alayrac2022flamingo} adopt perceiver resampler module to convert visual features into visual tokens and interleave the tokens in plain text at the locations of vision inputs. Besides, Flamingo  performed learnable cross attention to attend to the visual tokens of the image that appeared just before it in the interleaved sequence, rather than to all previous images and built a large-scale interleaved image-text dataset. On top of these design, recent works mainly focus on improving \textbf{instruction-following ability}. LLaMA-Adapters\cite{zhang2023llama} aims to adapt LLaMA\cite{touvron2023llama} into an instruction following model with an additional adapters module and multi-modal prompts. Mini-GPT4 \cite{zhu2023minigpt}, mPLUG-OWL \cite{ye2023mplug}, and InstructBLIP \cite{dai2023instructblip} adopt the Q-Former on various language models for instruction following capability. Besides, MultiModal-GPT\cite{gong2023multimodal} fine-tuned OpenFlamingo\cite{awadalla2023openflamingo} using Low-rank Adapter (LoRA)\cite{hu2021lora} and Otter\cite{li2023otter} introduced MultI-Modal In-Context Instruction Tuning (MIMIC-IT) dataset following three heuristics, both of which demonstrate improved instruction-following ability, Simultaneously.

\textbf{Grounded MLLM} Inspired by the success of MLLM, recent works focus on investigating the fine-grain grounding capability between region-text pairs instead of image-text pairs, and further conduct dense regional prediction tasks. 
One research line is to learn \textbf{regional alignment} between the image regions with the corresponding text tokens. RegionCLIP\cite{zhong2022regionclip} extends CLIP  with pseudo image
regions and textual concepts pairs. Grill\cite{jin2023grill} proposes to replace the referring words with the corresponding
visual patches to align text and image Regions. 
RegionBLIP\cite{zhou2023regionblip} takes position-assisted regional objects as soft prompts for LLM on image-region-text data.
Another research focus is to unleash the \textbf{grounding ability} in a multimodal large language model.
VL-T5\cite{cho2021unifying} converts the visual grounding task into regional \emph{ROI box} feature conditioned text generation to predict the box id. 
OFA\cite{wang2022ofa}, PEVL\cite{yao2022pevl} and 
KOSMOS-2 \cite{peng2023kosmos} reformulate continuous corner coordinates of object to \emph{discrete position tokens}. 
Shikra\cite{chen2023shikra} handles \emph{spatial coordinate} inputs and outputs in natural language without introducing extra vocabulary or position encoders. The works\cite{chen2023shikra,peng2023kosmos,zhang2023gpt4roi} also perform Instruction tuning and convert the position of regional
objects into language descriptions. Although gaze is flexible and interactive, it is easy for humans to understand the gaze's semantic representation but hard for AI agents.

\subsection{Region Representation for Large Language Models} 

The visual region can be represented as \textbf{bounding boxes}\cite{zhu2016visual7w,liu2017referring}, \textbf{points} \cite{mani2020point}, \textbf{traces}\cite{pont2020connecting,yan2021control}. Existing approaches usually leverage Fast-RCNN to detect bounding boxes which limits the pre-defined or recognized objects in the bounding box and hard to scale out. Points are flexible but are too fine-grain and require large amount of points to represent large regions precisely. Trace is a more natural way to input by using the mouse trace coordinates and most similar to human gaze. In AR and VR scenario, although trace is applicable with gesture, we propose to use gaze as a more convenient and interactive way. The two works\cite{chen2023shikra,mani2020point} takes bounding boxes or points as region input for visual question answering and are the most similar work. Different from them, we take \textbf{gaze} as regional inputs. 
 
\paragraph{Region Inputs}In order to input the regional information to the model, several methods\citep{zhang2023gpt4roi,bracha2023disclip} directly \textbf{concatenate cropped image patches} with the original text/image as model input. Another methods \citep{lin2020interactive,lin2022multi} use 0/1 \textbf{mask or Gaussian map} input with the original image to emphasize the area of user interest. Additionally, other methods \cite{kirillov2023segment,voigtlaender2023connecting} first encode points, boxes or trace to \textbf{positional encodings} then add them to intermediate features or learned queries. Specifically for gaze,  
\cite{qian2023gvgnet} propose a gaze-directed visual grounding and \emph{fuse} the gaze feature through a multi-modal fusion module. EG-ViT\citep{ma2023eye} propose a eye-gaze-guided
vision transformer which takes the \emph{masked image patches} within the gaze interest. 
{
\subsection{Gaze and Cursor as a proxy for Attention}
\label{rw-gazedata}
Cursor-based techniques, including approaches like SALICON\citep{jiang2015salicon} and BubbleView\citep{kim2017bubbleview}, have emerged as affordable, nonintrusive, and scalable alternatives to traditional eye-tracking methods for collecting human attentional data. Empirical evidence from prior work has established strong connections between cursor-like signals and gaze positions. Studies focusing on web browsing and search tasks have found a high correlation between cursor and gaze locations, with better alignment along the vertical dimension\citep{Huang2011NoCN,Huang2012UserSU,Guo2010TowardsPW,Chen2001WhatCA}. These findings support the motivation to use cursor-based techniques as a proxy for attention.

Despite their success, existing cursor-based studies have limitations, such as the need for complex post-processing of mouse movement data and evaluations limited to simple aggregate comparisons with eye-tracking data\citep{kim2017bubbleview}. Furthermore, while prior work serves as a solid foundation from a data-centric perspective, it lacks demonstration of whether modern applications aiming to assist users using gaze, such as vision language models (VLMs), can be trained from cursor data and later adapt to gaze signals, especially when transitioning from 2D planar images on screens to ego view scenes in head-mounted display (HMD) scenarios. Our work aims to directly tackle this problem, as we believe it is the optimal time to close the entire visionary loop of understanding and utilizing the gaze modality to ultimately achieve smart, in situ personal assistants.

\subsection{Saliency models on modeling gaze attention}
\label{rw-saliency}
\cite{Yarbus1967EyeMA} proposed that tasks could be decoded from fixation patterns, receiving mixed support in subsequent research. 
Early computational models of visual attention focused on bottom-up approaches, representing pre-attentive selection processes from visual input\citep{Koch1985ShiftsIS}. Later, the saliency map concept emerged\citep{Niebur1995ControlOS}.
Initially, models were trained on fixation data from eye-tracking experiments\citep{Kienzle2006ANA,Judd2009LearningTP}, but collecting large datasets proved difficult. The SALICON dataset\citep{jiang2015salicon} addressed this challenge by using mouse movements to simulate natural viewing behavior, leading to state-of-the-art performance in saliency models \citep{jiang2015salicon,Pan2016ShallowAD,Tavakoli2017SaliencyRA}. As deep learning advanced, saliency modeling improved\cite{Kmmerer2016DeepGazeIR},
enabling more complex gaze pattern modeling in vision-language tasks\citep{Sugano2016SeeingWH, Das2016HumanAI, Vasudevan2018ObjectRI}. 

\textbf{Saliency models aim to approximate the human visual system by predicting eye fixations on images} \citep{kim2017bubbleview}. Unlike traditional saliency models, \textbf{our approach takes ground truth gaze data, image, and natural language inputs to generate contextually relevant responses}, presenting a novel challenge. Recent work, such as \cite{Sugano2016SeeingWH}, leverages gaze signals to enhance captioning tasks but does not accommodate dynamic user queries beyond captioning. Additionally, their LSTM-based method falls short compared to contemporary large transformer baselines. \cite{sood21_conll} introduces gaze data to the visual question-answering (VQA) task, but their analysis remains limited to comparing human and neural attentive strategies learned by VQA models. 
With the development of large vision-language models, we believe our work provides a valuable contribution to modern applications by effectively tackling the new challenge.
}
\end{document}